\documentclass[10pt]{article} 
\usepackage[accepted]{tmlr}

\usepackage{amsmath,amsfonts,bm}

\def\eqref#1{equation~\ref{#1}}
\def\Eqref#1{Equation~\ref{#1}}

\def\1{\bm{1}}

\DeclareMathAlphabet{\mathsfit}{\encodingdefault}{\sfdefault}{m}{sl}
\SetMathAlphabet{\mathsfit}{bold}{\encodingdefault}{\sfdefault}{bx}{n}

\usepackage{algorithm}
\usepackage{algpseudocode}

\usepackage{hyperref}
\usepackage{url}

\usepackage{amssymb}
\usepackage{mathtools}
\usepackage{amsmath,amsthm}

\usepackage{graphicx}
\usepackage{adjustbox}
\usepackage{multirow}
\usepackage{subfigure}

\title{Universal and Efficient Detection of Adversarial Data through Nonuniform Impact on Network Layers}

\author{\name Furkan Mumcu \email furkan@usf.edu \\
      \addr Department of Electrical Engineering\\
      University of South Florida 
      \AND
      \name Yasin Yilmaz \email yasiny@usf.edu \\
      \addr Department of Electrical Engineering\\
      University of South Florida 
    }

\newcommand{\Input}[1]{\State \textbf{Input:} #1}
\newcommand{\Output}[1]{\State \textbf{Output:} #1}

\def\month{06}  
\def\year{2025} 
\def\openreview{\url{https://openreview.net/forum?id=0CY5APFnFI}} 

\begin{document}

\maketitle

\begin{abstract}

Deep Neural Networks (DNNs) are notoriously vulnerable to adversarial input designs with limited noise budgets. While numerous successful attacks with subtle modifications to original input have been proposed, defense techniques against these attacks are relatively understudied. Existing defense approaches either focus on improving DNN robustness by negating the effects of perturbations or use a secondary model to detect adversarial data. Although equally important, the attack detection approach, which is studied in this work, provides a more practical defense compared to the robustness approach. We show that the existing detection methods are either ineffective against the state-of-the-art attack techniques or computationally inefficient for real-time processing. We propose a novel universal and efficient method to detect adversarial examples by analyzing the varying degrees of impact of attacks on different DNN layers. {Our method trains a lightweight regression model that predicts deeper-layer features from early-layer features, and uses the prediction error to detect adversarial samples.} Through theoretical arguments and extensive experiments, we demonstrate that our detection method is highly effective, computationally efficient for real-time processing, compatible with any DNN architecture, and applicable across different domains, such as image, video, and audio.

\end{abstract}
\section{Introduction}

Deep neural networks (DNNs) are known to be vulnerable to subtle and manipulative noise for input data instances that can be designed by adversaries to cause erroneous outputs. 
Notably, \cite{fgsm} proposed a simple and effective method, called the Fast Gradient Sign Method (FGSM), to craft such adversarial instances by {adding or subtracting} a small perturbation to each input dimension {based on the sign of the gradient}. After FGSM, various adversarial sample generation methods were demonstrated across different domains. However, compared to the vast diversity among attack techniques, there {are} not enough {studies that} effectively and efficiently detect {the} increasing number of attacks in various domains.

There are two main defense strategies. The more ambitious one aims to mitigate the effects of attacks (e.g., correctly classifying adversarial images) by developing robust DNNs that are {increasingly less} vulnerable to adversarial data. As opposed to this idealistic objective, the more practical second strategy aims to detect and discard the adversarial data. This binary classification approach (natural vs. adversarial), which we use in this work, naturally provides a more tractable defense strategy than trying to solve the original DNN problem with adversarial data.

One of the oldest and most effective defense techniques is adversarial training for both the robustness \cite{fgsm} and detection \cite{grosse2017statistical}, in which the DNN is simply retrained using also the known adversarial instances. Although effective against known attacks, adversarial training suffers from high computational cost and unseen attacks. Similarly, other detection methods are typically only effective against some attacks. For instance, \cite{metzen2017detecting}, in a supervised learning setup, trains a binary classifier for attack detection, which {performs well on} the attacks seen in training, but fails for the unseen attacks. A recently proposed detector, EPS-AD \cite{eps-ad}, can detect unseen attacks by training only on natural data in a semi-supervised anomaly detection setup at the expense of significant computational cost. Another recent detection method, VLAD \cite{vlad}, can also detect a wide range of attacks by training a secondary model only on natural data, however it is vulnerable to transferable attacks that also affect its secondary baseline model. 


The existing adversarial data detectors in the literature are proposed for a particular application and do not naturally extend to other applications as they are based on domain-specific DNNs, e.g., adversarial image detection using convolutional neural networks \cite{metzen2017detecting} or diffusion models \cite{eps-ad}.
Motivated by the lack of a computationally efficient detector which can accurately detect a wide range of attacks in real-time, we propose a \emph{universal and efficient} detector, called Layer Regression (LR), that works together with DNNs in a variety of applications, including image recognition, video action recognition, and speech recognition. Our contributions can be summarized as follows:

\begin{itemize}
    \item We propose the first universal and efficient adversarial data detection method, which takes advantage of the nonuniform impacts of adversarial samples on different DNN layers.
    
    \item We conduct extensive experiments for image recognition, and show that LR significantly outperforms existing efficient methods. 
    Universality of LR is shown with its superior performance in detecting action recognition attacks and speech recognition attacks. 
    \item In addition to its high performance across a wide range of domains, models, and attacks, LR is also the most lightweight defense method. It is orders of magnitude faster than the existing defenses, making it ideal for real-time attack detection and resource-constrained systems.
\end{itemize}

\section{Related Work}

\subsection{Adversarial Attacks}

The robustness of deep neural networks and their vulnerability {to} adversarial examples have been investigated since the introduction of FGSM \citep{fgsm}. Numerous adversarial attacks have been proposed to generate effective adversarial examples in recent years \citep{pgd, apgd, bim, chen2017zoo, ilyas2018black, mumcu2024sequential, vmi_vni, anda, pif}. There are two main adversarial attack settings, namely white-box and black-box. While it is assumed that the attacker has access to the target model in the white-box setting, in the black-box setting, the attacker does not have any prior information about the target model. 

White box attacks, including FGSM \citep{fgsm}, PGD \citep{pgd}, APGD \citep{apgd}, generate adversarial examples by maximizing the target model's loss function and they are usually referred {to} as gradient-based attacks. BIM \citep{bim} tries to improve gradient-based attack by applying perturbations iteratively. 
Transferability based black-box attacks, which {were} introduced in \cite{papernot2017practical}, {among} one of the most common black-box approach{es}. The idea is to use an attack on a substitute model for generating adversarial examples for unknown target models, utilizing the transferability of adversarial examples to different DNNs. While adversarial examples generated by these attacks are most effective when the substitute model exactly matches the target model as in a white-box attack setting, their success is shown to be transferable even when there is significant architectural differences between the substitute and target models.
\cite{vmi_vni} introduced VMI and VNI to further extend iterative  gradient-based attacks and try to achieve high transferability by considering the gradient variance of the previous iterations. PIF \citep{pif} uses patch-wise iterations to achieve transferability. ANDA \citep{anda} aims to achieve strong transferability by avoiding the overfitting of adversarial examples to the substitute model.
In addition, there are approaches that combine multiple attacks, such as AutoAttack \citep{aa}, to test the robustness of models against a diverse set of adversarial perturbations.

\subsection{Adversarial Defenses}
\label{defenses}

Attempting to make changes on the input data for removing the effects of perturbations from adversarial examples is the most common defense strategy.
JPEG compression is studied in several works \citep{jpeg1, jpeg2, jpeg3}, and {it has been shown that} compressing and decompressing helps to remove adversarial effects {from} input images. \cite{random} uses random resizing and padding  on the inputs to eliminate the adversarial effects.
Several denoising methods \citep{denoise1, denoise2, denoise3} were proposed to remove adversarial perturbations from the inputs.
\cite{wdsr} uses wavelet denoising and image super resolution as pre-processing steps to create a defense pipeline against adversarial attacks.
\cite{deflection} tries to eliminate adversarial effects by redistributing the pixel values via a process called pixel deflection. Adversarial training is another method which is studied to increase robustness of DNNs, however adversarial training often fails to perform well especially under various attack configurations \citep{adv_training}.
\cite{papernot2018deep} proposes to increase the robustness of the model by applying k-nearest neighbors (kNN) classification to the feature representations at different layers of a deep neural network. This method also implicitly utilizes the nonuniform impact of adversarial data on network layers, however their deep kNN method is not universally applicable to any deep neural network and not computationally efficient, unlike our proposed LR method. 
On the other hand, several detection methods were proposed in recent years. 
\cite{squeezing} introduces feature squeezing where they use bit reduction, spatial smoothing, and non-local means denoising to detect adversarial examples. \cite{eps-ad} tries to {detect} adversarial samples by computing an Expected Perturbation Score (EPS), which averages a sample’s behavior over multiple perturbations generated using a pre-trained diffusion model. \cite{contra} detects adversarial examples by identifying semantic contradictions by using a generator that reconstructs the input based on the network’s internal feature representations. 
{\cite{pang2018rce} proposed a training-time modification using reverse cross-entropy to enforce separable feature representations for clean and adversarial inputs, which improves detection performance but requires modifying the training process. Similarly, \cite{tian2018detecting} introduced a detection method based on prediction consistency under image transformations, leveraging the instability of adversarial samples under rotation or translation, whereas our method operates on internal feature inconsistencies without needing input perturbations or model retraining.}
A recent defense strategy which utilizes a baseline model, e.g., Vision Language Models (VLMs), with the assumption that the output of target model and baseline model are close to each other for clean input, but are far away from each other for adversarital input. A recent example is demonstrated by \cite{vlad} where they used CLIP \citep{clip} to detect adversarial video examples.


\section{Detecting Adversarial Examples}

Consider a Deep Neural Network (DNN) model $g(\cdot)$ that takes an input $x$ and predicts the target variable $y$ with $g(x)$.
As discussed in Section \ref{defenses}, there are three main defense strategies against adversarial attacks: adversarial training, modifying input, and detecting adversarial samples by monitoring changes in output with respect to a baseline. 
While the former two focuses on the changes in the input ($x$ vs. $x^{adv})$, the latter utilizes the changes in the output ($g(x)$ vs. $g(x^{adv}))$. Our approach differs from these existing approaches by leveraging the nonuniform changes among different DNN layer activations. Instead of analyzing the input $x^{adv}$ or the output $g(x^{adv})$, the proposed defense method analyzes the intermediate steps between $x^{adv}$ and $g(x^{adv})$.

To develop a universal detector that work{s} with any DNN and against any attack, we start with the following generic observation. 
Although there are numerous attacks with different approaches to generate adversarial examples, all attacks essentially aim to change the model's prediction by maximizing the loss $\mathcal{L}$ (e.g., cross-entropy loss) between prediction $g(x^{adv})$ and the {one-hot encoded} ground truth $y$ while limiting the perturbation \cite{anda}:
\begin{align}  
\label{eq:min}
\max_{x^{adv}} \mathcal{L}(g(x^{adv}),y)  ~~\text{s.t.}~~ \|x^{adv} - x\|_{\infty} \leq \epsilon.
\end{align}

Considering this common aim of attack methods, "start with a small perturbation and end up with a big one", and the sequential nature of DNNs, we hypothesize that the impact of adversarial examples on the final layer is higher than the initial layer. Let us first define a generic DNN $g(\cdot)$ consisting of $n$ layers $a=\{a_1, a_2, ..., a_n\}$. In DNNs, including CNNs, transformers, etc., layers incrementally process the information from the previous layers to compute their respective outputs to the next layer. For example, for a model $g$ where each layer is connected to the previous one, the final output of the model can be formulated as follows:
\begin{align}  
\label{eq:dnn}
g(x) = a_n(a_{n-1}(\dots a_1(x))). 
\end{align}
For simplicity, we will denote a layer's output vector with $a_i(x)$. Note that, the output of {the} last layer $a_n(x)=g(x)$ is typically the class probability vector in classification {tasks}, and $a_{n-1}(x)$ is referred as the feature vector of the model. 

{Our Conjecture and Remark below to build a universal and efficient detector are based on the assumption that, for an adversarial sample $x^{adv}$, the first layer output $a_1(x^{adv})$ remains close to the clean version $a_1(x)$ since the change in the input is required to be unnoticeable by design, i.e., $\|x^{adv} - x\|_{\infty} \leq \epsilon$.}

\textbf{\textit{Conjecture:}}
    Assume a loss function $\mathcal{L}(g(x),y)$ that is monotonic with $\|g(x)-y\|_\infty$. 
    {Assuming a highly accurate target model $g$ that is trained by optimizing the weights $w$ to effectively minimize the loss $\mathcal{L}(g_w(x),y)$, i.e., 
    \begin{equation}
        g(x) = g_{w^*}(x), ~\text{where}~ w^* = \arg\min_w \mathcal{L}(g_w(x),y),
    \end{equation}}
    we can rewrite  \Eqref{eq:min} as 
    \begin{equation}  
        \max_{x^{adv}} \mathcal{L}(g(x^{adv}),g(x))  ~~\text{s.t.}~~ \|x^{adv} - x\|_{\infty} \leq \epsilon.
    \end{equation}
    Note that $a_n(x)=g(x)$, $a_n(x^{adv})=g(x^{adv})$, and $\mathcal{L}(g(x^{adv}),g(x))$ is monotonic with $\|g(x^{adv})-g(x)\|_\infty$. Hence, 
    $\|a_n(x^{adv})-a_n(x)\|_\infty$ is maximized while limiting $\|x^{adv} - x\|_{\infty}$ by a small number. 
    
    {\emph{Since the first layer is right next to the input, for which the perturbation is minimized, and far away from the final layer, for which the impact is maximized, we conjecture that, for $n>1$,}} 
    \begin{equation}
    \label{eq:thm}
        d_1 = \|a_1(x^{adv})-a_1(x)\|_\infty < d_n = \|a_n(x^{adv})-a_n(x)\|_\infty.
    \end{equation}

According to this conjecture, the impact of an adversarial sample is higher on the final layer output than the first layer output. We next present how to utilize this for detecting adversarial samples. 

\textbf{\textit{Remark:}}
    Consider a function $f$ which maps $a_1(x)$ to $a_n(x)$ and is stable in the sense that 
    $\|f(a_1(x))-f(a_1(x+\epsilon))\|_\infty \leq \delta$
    for small $\epsilon$ and $\delta$. 
    Since $\|x^{adv}-x\|_\infty\leq \epsilon$ from  \Eqref{eq:min}, {assuming $a_1(x^{adv})$ stays close to $a_1(x)$,} $f(a_1(x^{adv}))$ is close to $f(a_1(x))$ due to the stability of $f$. From  \Eqref{eq:thm}, $a_n(x^{adv})$ is far away from $a_n(x)$ compared to the distance between $a_1(x)$ and $a_1(x^{adv})$. Thus, the estimation error for adversarial samples is larger than the error for clean samples:
    \begin{equation}
        \label{eq:cor}
        e_a = \|f(a_1(x^{adv}))-a_n(x^{adv})\|_\infty > e_c = \|f(a_1(x))-a_n(x)\|_\infty.
    \end{equation}

\subsection{Layer Regression} 
\label{sec:lr}

\begin{figure*}[t]
    \centering
    \includegraphics[width=0.4\linewidth]{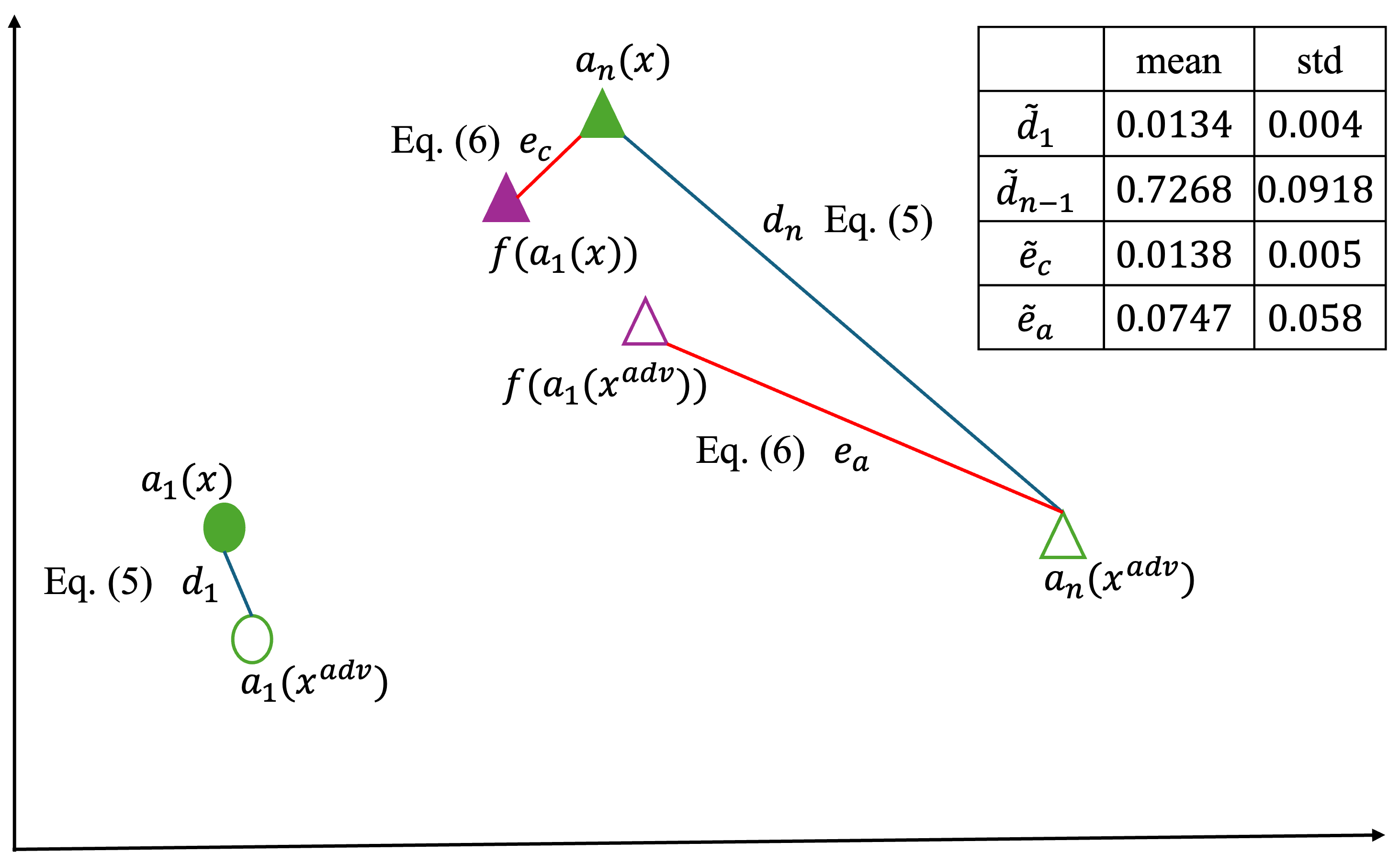}
    \includegraphics[width=0.4\linewidth]{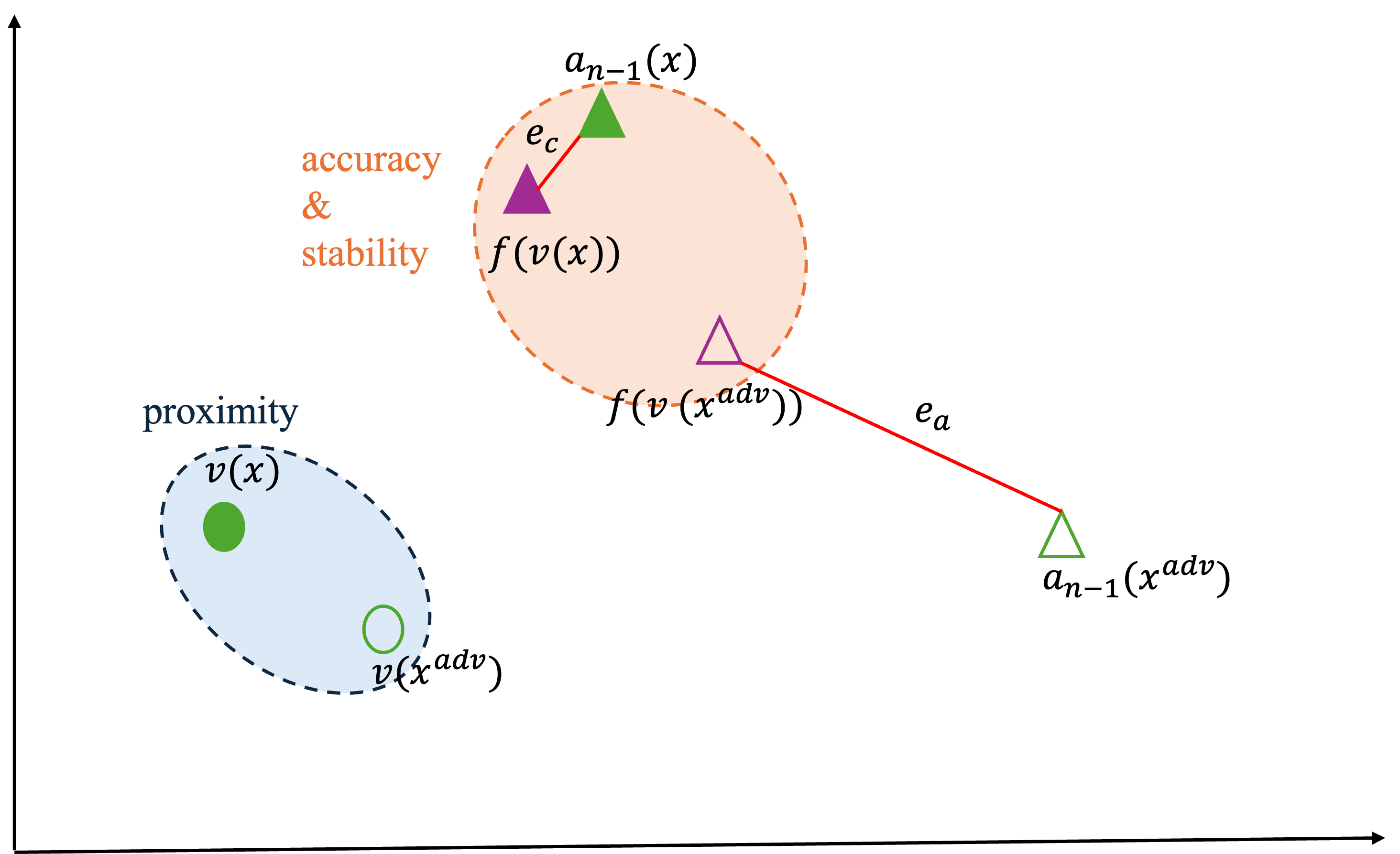}
    \caption{(Left)  \Eqref{eq:thm} conjectures that the impact of adversarial samples is higher on the final layer than the first layer.  \Eqref{eq:cor} uses this to show that the error of a stable estimator is higher for adversarial samples compared to clean samples. (Right) The performance of approximations in proposed detector to  \Eqref{eq:cor} depend on two conflicting objectives: proximity of input vectors for clean and adversarial samples, and training an accurate and stable estimator. (Table) Empirical confirmation of the agreement between the proposed detector with approximations and the theoretical arguments.}
    \label{fig:v_sel}
\end{figure*}

While  \Eqref{eq:thm} provides the theoretical motivation, the result in  \Eqref{eq:cor} {offers} a mechanism to detect adversarial samples if we can train a suitable function $f$ (Figure \ref{fig:v_sel}). We make four approximations to obtain a practical algorithm based on  \Eqref{eq:cor}. First, for computational efficiency and real-time detection even in resource-constrained systems, we propose to use a multi-layer perceptron (MLP) to approximate $f$. Second, since $a_n(x)$ denotes the predicted class probabilities, we choose the feature vector $a_{n-1}$, which takes unconstrained real values, as the target to train MLP as a regression model. Third, we approximate non-differentiable $\|\cdot\|_\infty$ with $\|\cdot\|_2$ to train the MLP using the differentiable mean squared error (MSE) loss. 
To empirically {validate} of  \Eqref{eq:thm} under these three approximations, we use the ImageNet validation dataset, ResNet-50 \cite{resnet} as target model, and PGD attack \cite{pgd} to compute the mean and standard deviation of normalized change in layer 1, $\tilde{d}_1=\frac{\|a_1(x^{adv})-a_1(x)\|_2}{\|a_1(x^{adv})\|_2+\|a_1(x)\|_2}$, and layer $n-1$, $\tilde{d}_{n-1}=\frac{\|a_{n-1}(x^{adv})-a_{n-1}(x)\|_2}{\|a_{n-1}(x^{adv})\|_2+\|a_{n-1}(x)\|_2}$.

\begin{figure*}[ht]
    \centering
    \includegraphics[width=1\linewidth]{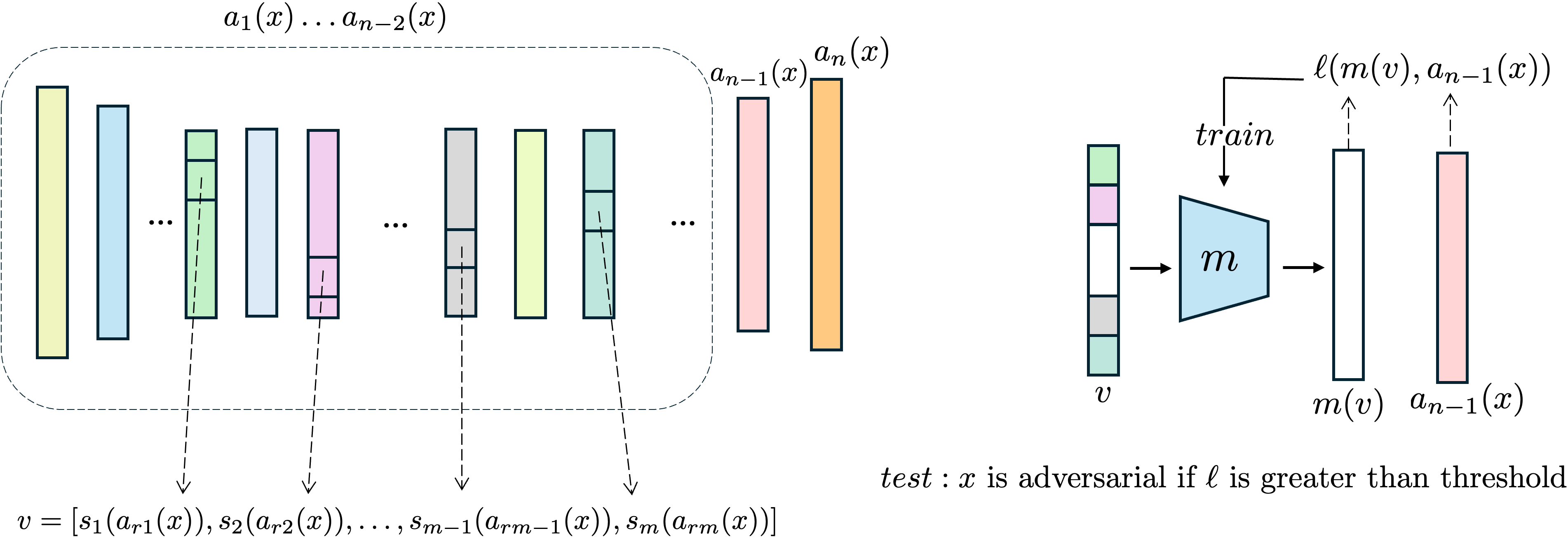}
    \caption{(Left) Layer selection and slicing operations to form the input vector. (Right) Training and testing procedures of the proposed detector.}
    \label{fig:method}
\end{figure*}

In deep neural networks with $n\gg 1$, training {a} suitable $f$ to estimate the feature vector $a_{n-1}$ using the first layer output $a_1$ as the input is a challenging task due to the highly nonlinear mapping in $n-2$ layers. To develop a lightweight detector via MLP, as the fourth approximation, we propose selecting a mixture of early layer outputs as the input to the regression model instead of using only the first layer. Using a mixture of 5th, 8th, and 13th convolutional layers in ResNet-50 as the input, we empirically check  \Eqref{eq:cor} under the same setting used for $\tilde{d}_1$ and $\tilde{d}_{n-1}$ by computing the mean and standard deviation of MSE $\tilde{e}_c = \|f(a_1(x))-a_{n-1}(x)\|_2$ for clean images and $\tilde{e}_a = \|f(a_1(x^{adv}))-a_{n-1}(x^{adv})\|_2$ for adversarial images, where an MLP with two hidden layers is used for $f$. Results shown in Figure \ref{fig:v_sel} corroborate  \Eqref{eq:thm} and  \Eqref{eq:cor} under the approximations. Input selection for MLP is further discussed in this section and in {Supplementary \ref{ap:inpu_sel}.}

Utilizing the four approximations to  \Eqref{eq:cor} discussed above, we propose a universal and lightweight detection algorithm, as shown in Figure \ref{fig:method}, with the following steps: (i) select a subset of the first $n-2$ layer vectors and generate a new vector $v$ from the selected subset, (ii) feed $v$ into a regression model $m$ to predict the feature vector $a_{n-1}(x)$, (iii) train $m$ by minimizing the mean squared error (MSE) loss $\ell(m(v),a_{n-1}(x))$ in the clean training set devoid of adversarial samples, 
(iv) use $\ell(m(v),a_{n-1}(x))$ as the detection score. $m$ is expected to produce low scores for clean inputs and high for adversarial inputs. 
A pseudo-code of LR is given in Supplementary \ref{ap:pseudo}.


The formation of vector $v$ can be formed in various ways, such as using only the $i$th layer vector $v=a_i(x)$ or mixture of several layer vectors. To enable larger estimation error $e_a$ for adversarial samples than estimation error $e_c$ for clean samples, the choice for $v$ should strike a balance between two competing goals, as illustrated in Figure \ref{fig:v_sel}: proximity of clean $v(x)$ and adversarial $v(x^{adv})$, and accuracy and stability of estimation function $f$. While training an accurate and stable regression model is more feasible when $v$ is selected from the layers closer to the target layer $n-1$, e.g., $v=a_{n-2}(x)$, such a detector might be less sensitive to adversarial samples since both $a_{n-2}(x)$ and $a_{n-1}(x)$ are expected to be impacted significantly by the attack, i.e., $a_{n-2}(x)$ and $a_{n-2}(x^{adv})$ will not be proximal. On the other hand, while selecting $v=a_1(x)$ ensures a reasonably small perturbation in $v$, it also makes obtaining an accurate and stable estimator more challenging. As a result, we propose to select a subset of layer vectors 
\begin{equation}
    a_r = \{a_{r1}(x), a_{r2}(x), ..., a_{rm}(x)\}
    \label{eq:ar}
\end{equation} 
where $a_r \in a$ and $m < n$ is the number of selected layers. From the selected layer vectors, we aim to generate a new vector $v$. However, since the layer vectors are often large due to the operations like convolutions or attentions, to get a specific portion of the selected layer vectors, we define a unique slicing function  $s = \{s_1, s_2, ..., s_m\}$ for each layer vector in $a_r$. Then, each slicing function is applied to the corresponding layer vector in $a_r$ to get the sliced vectors  
\begin{equation}
    s_r = \{s_1(a_{r1}(x)), s_2(a_{r2}(x)), ..., s_m(a_{rm}(x))\}.
    \label{eq:slicing}
\end{equation}
Finally, the vector $v$ is generated by concatenating the vectors in $s_r$: 
\begin{align}  
\label{eq:v_vector}
v = [s_1(a_{r1}(x)), s_2(a_{r2}(x)), \ldots, s_m(a_{rm}(x))].
\end{align}
The order of $s_i(a_{ri}(x))$ in $v$ can be randomized in inference to counteract adaptive attacks, as explained in Section \ref{sec:aa}. 
The proposed layer selection and slicing process is summarized in Figure \ref{fig:method}. During the training, only the clean input samples are used. After the training, the loss is expected to be low for clean inputs and high for adversarial inputs. The implementation of Layer Regression is publicly available at \url{https://github.com/furkanmumcu/Layer-Regression}.


\section{Experiments}
\label{sec:exps}


\textbf{Datasets and Evaluation:} For the evaluation on ImageNet, 10,000 images from the validation set were chosen. For CIFAR-100, the test set, which also contains 10,000 images, was used.
The commonly used area under the receiver operating characteristic (AUROC) curve is used to evaluate the attack detection performance of defense methods. 
Similarly to existing work \citep{eps-ad}, the evaluation {is} conducted on adversarial and clean sets.
The clean set comprises all images that are correctly classified by the target models. An adversarial set is formed for each attack-target model combination by gathering the attack's adversarial images misclassified by target model.
More details on the evaluation methods and datasets are given in Supplementary \ref{ap:subsets}.


\textbf{Baselines:} We benchmark our {method} against a diverse range of baseline approaches, including six popular computationally efficient methods: JPEG compression (JPEG) \citep{jpeg3}, Randomization (Random)\citep{random}, Deflection (Deflect) \citep{deflection}, Feature Squeezing (FS) \cite{fs}, Wavelet denoising (Denoise) and super resolution (WDSR) \citep{wdsr},
and two most recent state-of-the-art detection methods: VLAD \citep{vlad} and EPS-AD \citep{eps-ad}. Since FS, VLAD and EPS-AD were proposed as detectors in their respective papers, we use their official implementations. The remaining methods were proposed to increase robustness by altering the inputs to remove perturbations. {Therefore, we derive} a detection method from them by comparing the predictions {before and after} the methods are applied. The case the predictions match indicates no attack and a mismatch indicates an attack.


\textbf{Target Models}: The experiments are conducted using the validation set of ImageNet dataset \citep{imagenet} and the CIFAR-100 dataset \citep{cifar}. 
To represent the different architectures we used three CNN-based image classification models, VGG19 \cite{vgg}, ResNet50 \cite{resnet},  InceptionV3 \cite{inception}; and three {transformer-}based models, ViT \cite{vit}, DeiT \cite{deit}, LeViT \cite{levit}.

\textbf{Attack Methods:}
In this section, we use the untargeted $l_\infty$ attack setting, however we also report our detector's performance under targeted and $l_2$ attack settings in Supplementary \ref{ap:tab:targeted-l2}. 
We consider two different threat models in the experiments. The first one is the white-box static attack setting {in which} the attacker has complete knowledge of the classifier but not the detector. Strong attacks from the literature, namely the 
BIM \citep{bim}, PGD \citep{pgd}, PIF \citep{pif}, APGD \citep{apgd}, ANDA \citep{anda}, VMI and VNI \citep{vmi_vni} attacks are used as white-box attacks. We also test against an ensemble white-box attack, called AutoAttack (AA) \citep{aa}. 

The second threat model is the adaptive attack setting in which the attacker also has the full knowledge of defense mechanism. We analyze the performance of our detector against a specially crafted adaptive attack \cite{contra} which is trained to simultaneously deceive the target model and {bypass} our detector in Section \ref{sec:aa}.



\textbf{LR training:} For each model, an MLP with 2 hidden layers is trained as the LR detector {to demonstrate that a highly effective detector can be built using a very lightweight neural network with minimal computational overhead}.  After experimenting with different layer selection and slicing strategies, we found that selecting only from early or final layers reduce the performance, but randomly selecting three layers and slicing the middle 60\% portion of each selected layer yields good results. 
The selection procedure of layers $a_r$ and slicing functions $s_r$, and training parameters are detailed in Supplementary \ref{ap:subsets}.


\subsection{Detecting White-Box Static Attacks}
\label{sec:white}

\begin{table*}[t]
\Large
\begin{adjustbox}
{width=.85\linewidth,center}
  \centering
  \begin{tabular}{c  c | c  c  c  c  c c  c |  c  c  c  c c c  c }
     &  & \multicolumn{7}{c|}{\textbf{ImageNet}} & \multicolumn{7}{c}{\textbf{CIFAR-100}} \\

     &  & JPEG  & Random & Deflect  & Denoise & WDSR & FS  & LR (Ours)  & JPEG  & Random & Deflect  & Denoise & WDSR & FS  & LR (Ours)\\


    \hline

            \multirow{7}{*}{\rotatebox[origin=c]{90}{VGG19}}
& BIM & 0.42 &  0.43 &  0.48 &  0.45 &  0.39 &  0.13 &   \textbf{0.99}  & 0.47 &  0.31 &  0.49 &  0.48 &  0.41 &  0.01 &  \textbf{0.99} \\
& PGD & 0.50 &  0.46 &  0.48 &  0.45 &  0.46 &  0.22 &   \textbf{0.99}  & 0.48 &  0.31 &  0.49 &  0.48 &  0.44 &  0.02 &   \textbf{0.99} \\
& PIF & 0.33 &  0.44 &  0.48 &  0.45 &  0.31 &  0.08 &   \textbf{0.99}  & 0.47 &  0.31 &  0.49 &  0.48 &  0.41 &  0.01 &   \textbf{0.99} \\
& APGD & 0.53 &  0.46 &  0.48 &  0.45 &  0.49 &  0.23 &   \textbf{0.99}  & 0.48 &  0.32 &  0.49 &  0.48 &  0.42 &  0.02 &   \textbf{0.99} \\
& ANDA & 0.39 &  0.47 &  0.49 &  0.47 &  0.38 &  0.26 &   \textbf{0.95}  & 0.48 &  0.36 &  0.49 &  0.49 &  0.43 &  0.11 &   \textbf{0.99} \\
& VMI & 0.36 &  0.42 &  0.48 &  0.45 &  0.34 &  0.11 &   \textbf{0.99}  & 0.47 &  0.32 &  0.49 &  0.48 &  0.41 &  0.01 &   \textbf{0.99} \\
& VNI & 0.41 &  0.45 &  0.48 &  0.46 &  0.38 &  0.19 &   \textbf{0.99}  & 0.47 &  0.32 &  0.49 &  0.48 &  0.42 &  0.02 &   \textbf{0.99} \\
  \hline 

           \multirow{7}{*}{\rotatebox[origin=c]{90}{ResNet50}}
& BIM & 0.67 &  0.63 &  0.49 &  0.48 &  0.55 &  0.29 &   \textbf{0.99}  & 0.82 &  0.69 &  0.46 &  0.62 &  0.73 &  0.39 &   \textbf{0.98} \\
& PGD & 0.77 &  0.75 &  0.49 &  0.49 &  0.68 &  0.53 &   \textbf{0.98}  & 0.87 &  0.71 &  0.45 &  0.71 &  0.82 &  0.40 &   \textbf{0.99} \\
& PIF & 0.51 &  0.64 &  0.49 &  0.48 &  0.47 &  0.29 &   \textbf{0.96}  & 0.73 &  0.72 &  0.45 &  0.48 &  0.61 &  0.63 &   \textbf{0.97} \\
& APGD & 0.75 &  0.70 &  0.49 &  0.49 &  0.64 &  0.47 &   \textbf{0.97}  & 0.84 &  0.65 &  0.46 &  0.62 &  0.78 &  0.44 &   \textbf{0.96} \\
& ANDA & 0.45 &  0.48 &  0.49 &  0.48 &  0.44 &  0.14 &   \textbf{0.96}  & 0.56 &  0.53 &  0.45 &  0.48 &  0.45 &  0.07 &   \textbf{0.99} \\
& VMI & 0.49 &  0.52 &  0.49 &  0.48 &  0.45 &  0.14 &   \textbf{0.99}  & 0.61 &  0.58 &  0.45 &  0.50 &  0.50 &  0.23 &   \textbf{0.99} \\
& VNI & 0.53 &  0.54 &  0.49 &  0.49 &  0.47 &  0.21 &   \textbf{0.97}  & 0.62 &  0.55 &  0.45 &  0.50 &  0.51 &  0.28 &   \textbf{0.97} \\

  \hline
            \multirow{7}{*}{\rotatebox[origin=c]{90}{InceptionV3}}
& BIM & 0.55 &  0.53 &  0.50 &  0.49 &  0.77 &  0.24 &   \textbf{0.98}  & 0.88 &  0.80 &  0.43 &  0.74 &  0.81 &  0.36 &   \textbf{1.00} \\
& PGD & 0.62 &  0.58 &  0.50 &  0.49 &  0.80 &  0.36 &   \textbf{0.97}  & 0.86 &  0.66 &  0.43 &  0.76 &  0.81 &  0.32 &   \textbf{0.99} \\
& PIF & 0.48 &  0.52 &  0.50 &  0.49 &  0.67 &  0.11 &   \textbf{0.99}  & 0.82 &  0.89 &  0.44 &  0.57 &  0.72 &  0.34 &   \textbf{1.00} \\
& APGD & 0.60 &  0.56 &  0.50 &  0.50 &  0.79 &  0.36 &   \textbf{0.96}  & 0.88 &  0.82 &  0.43 &  0.77 &  0.81 &  0.38 &   \textbf{0.99} \\
& ANDA & 0.49 &  0.50 &  0.50 &  0.49 &  0.52 &  0.23 &   \textbf{0.92}  & 0.71 &  0.57 &  0.43 &  0.50 &  0.59 &  0.29 &   \textbf{0.99} \\
& VMI & 0.48 &  0.50 &  0.50 &  0.49 &  0.67 &  0.12 &   \textbf{0.98}  & 0.85 &  0.78 &  0.42 &  0.50 &  0.80 &  0.35 &   \textbf{1.00} \\
& VNI & 0.52 &  0.53 &  0.50 &  0.50 &  0.70 &  0.25 &   \textbf{0.96}  & 0.86 &  0.79 &  0.43 &  0.60 &  0.81 &  0.31 &   \textbf{1.00} \\
  \hline
            \multirow{7}{*}{\rotatebox[origin=c]{90}{ViT}}
& BIM & 0.87 &  0.90 &  0.53 &  0.64 &  0.82 &  0.93 &   \textbf{0.99}  & 0.50 &  0.66 &  0.50 &  0.50 &  0.49 &  0.29 &   \textbf{0.98} \\
& PGD & 0.85 &  0.87 &  0.53 &  0.60 &  0.79 &  0.91 &   \textbf{0.99}  & 0.49 &  0.67 &  0.50 &  0.50 &  0.48 &  0.27 &   \textbf{0.97} \\
& PIF & 0.79 &  0.82 &  0.52 &  0.51 &  0.67 &  0.87 &   \textbf{0.99}  & 0.49 &  0.53 &  0.50 &  0.49 &  0.48 &  0.15 &   \textbf{0.98} \\
& APGD & 0.86 &  0.90 &  0.53 &  0.63 &  0.80 &  0.92 &   \textbf{0.99}  & 0.50 &  0.71 &  0.50 &  0.50 &  0.50 &  0.32 &   \textbf{0.95} \\
& ANDA & 0.72 &  0.68 &  0.52 &  0.56 &  0.64 &  0.82 &   \textbf{0.97}  & 0.49 &  0.52 &  0.50 &  0.50 &  0.48 &  0.26 &   \textbf{0.85} \\
& VMI & 0.77 &  0.81 &  0.51 &  0.57 &  0.70 &  0.86 &   \textbf{0.99}  & 0.49 &  0.58 &  0.50 &  0.49 &  0.48 &  0.15 &   \textbf{0.99} \\
& VNI & 0.78 &  0.83 &  0.52 &  0.61 &  0.73 &  0.91 &   \textbf{0.99}  & 0.50 &  0.60 &  0.50 &  0.50 &  0.49 &  0.27 &   \textbf{0.95} \\

  \hline
            \multirow{7}{*}{\rotatebox[origin=c]{90}{DeiT}}
& BIM & 0.86 &  0.89 &  0.52 &  0.58 &  0.78 &  0.88 &   \textbf{0.99}  & 0.50 &  0.64 &  0.50 &  0.49 &  0.48 &  0.36 &   \textbf{0.92} \\
& PGD & 0.86 &  0.88 &  0.53 &  0.58 &  0.80 &  0.90 &   \textbf{0.99}  & 0.50 &  0.66 &  0.50 &  0.49 &  0.48 &  0.34 &   \textbf{0.87} \\
& PIF & 0.77 &  0.84 &  0.51 &  0.50 &  0.71 &  0.62 &   \textbf{0.99}  & 0.48 &  0.52 &  0.50 &  0.49 &  0.47 &  0.16 &   \textbf{0.97} \\
& APGD & 0.85 &  0.90 &  0.52 &  0.57 &  0.78 &  0.88 &   \textbf{0.99}  & 0.50 &  0.65 &  0.50 &  0.50 &  0.48 &  0.36 &   \textbf{0.91} \\
& ANDA & 0.77 &  0.76 &  0.53 &  0.58 &  0.70 &  0.75 &  \textbf{0.99}  & 0.49 &  0.51 &  0.50 &  0.49 &  0.48 &  0.26 &   \textbf{0.91} \\
& VMI & 0.79 &  0.84 &  0.51 &  0.53 &  0.69 &  0.80 &   \textbf{0.99}  & 0.49 &  0.56 &  0.50 &  0.49 &  0.48 &  0.30 &   \textbf{0.96} \\
& VNI & 0.80 &  0.85 &  0.52 &  0.55 &  0.72 &  0.85 &   \textbf{0.99}  & 0.49 &  0.57 &  0.50 &  0.50 &  0.48 &  0.28 &   \textbf{0.94} \\

  \hline
            \multirow{7}{*}{\rotatebox[origin=c]{90}{LeViT}}
& BIM & 0.68 &  0.73 &  0.50 &  0.50 &  0.62 &  0.64 &   \textbf{0.99}  & 0.93 &  0.83 &  0.68 &  0.88 &  0.86 &  0.98 &   \textbf{0.99} \\
& PGD & 0.69 &  0.74 &  0.50 &  0.49 &  0.62 &  0.66 &   \textbf{0.99}  & 0.93 &  0.84 &  0.57 &  0.84 &  0.85 &  0.86 &   \textbf{0.93} \\
& PIF & 0.54 &  0.65 &  0.50 &  0.49 &  0.49 &  0.51 &   \textbf{0.99}  & 0.86 &  0.82 &  0.50 &  0.53 &  0.80 &  0.62 &   \textbf{0.93} \\
& APGD & 0.75 &  0.80 &  0.50 &  0.51 &  0.69 &  0.76 &   \textbf{0.98}  & 0.93 &  0.83 &  0.68 &  0.88 &  0.86 &  0.98 &   \textbf{0.99} \\
& ANDA & 0.49 &  0.51 &  0.50 &  0.49 &  0.49 &  0.43 &   \textbf{0.94}  & 0.84 &  0.71 &  0.52 &  0.57 &  0.81 &  0.65 &   \textbf{0.93} \\
& VMI & 0.53 &  0.59 &  0.50 &  0.49 &  0.48 &  0.39 &   \textbf{0.99}  & 0.93 &  0.83 &  0.66 &  0.66 &  0.86 &  0.98 &   \textbf{0.99} \\
& VNI & 0.60 &  0.66 &  0.50 &  0.51 &  0.54 &  0.59 &   \textbf{0.99}  & 0.92 &  0.83 &  0.67 &  0.71 &  0.86 &  0.99 &   \textbf{0.99}\\

  \hline
& Average & 0.63 &  0.66 &  0.50 &  0.51 &  0.61 &  0.50 &   \textbf{0.99}  & 0.65 &  0.62 &  0.50 &  0.56 &  0.60 &  0.35 &   \textbf{0.97} \\


  \end{tabular}
  \end{adjustbox}
  \caption{Comparison between 7 computationally efficient defenses in terms of detection AUROC against 7 attack methods targeting 6 models using ImageNet and CIFAR-100 datasets.}
  \label{tab:main-results}
\end{table*}

\begin{table}[t]
  \small
  \centering
  \begin{adjustbox}{width=.55\columnwidth,center}
  \begin{tabular}{c  c | c  c  c  c c c c }
    
    &  & BIM & PGD & PIF & APGD & ANDA & VMI & VNI  \\
    \hline

            \multirow{3}{*}{\rotatebox[origin=c]{90}{VGG19}}
            &VLAD        & 0.94 & 0.94	& 0.96	& 0.96 & 0.93 & 0.93	& 0.95\\
		&EPS-AD      &  \textbf{0.99} & 0.97 & 0.97	&  \textbf{0.99} & 0.98 &  \textbf{0.99}	&  \textbf{0.99}\\
		&LR (Ours)  &  \textbf{0.99} &  \textbf{0.99} &  \textbf{0.99}	&  \textbf{0.99} &  \textbf{0.99} &  \textbf{0.99}	&  \textbf{0.99}\\
  \hline 

            \multirow{3}{*}{\rotatebox[origin=c]{90}{ResNet}}
            &VLAD       & 0.81   & 0.81	   & 0.81	& 0.82 & 0.77 & 0.80	& 0.80\\
		&EPS-AD     & \textbf{0.99}    & 0.98   &  \textbf{0.98}	&  \textbf{0.99} &  \textbf{0.99} &  \textbf{0.99}	&  \textbf{0.99}\\
		&LR (Ours)  & \textbf{0.99}    &  \textbf{0.99}   &  \textbf{0.98}	& 0.97 & 0.96 &  \textbf{0.99}	& 0.97\\
\hline
            \multirow{3}{*}{\rotatebox[origin=c]{90}{ViT}}
            &VLAD  & 0.94 & 0.92	& 0.89	& 0.93 & 0.85 & 0.93	& 0.91\\
		&EPS-AD  & \textbf{0.99} & 0.97 & 0.98	&  \textbf{0.99} &  \textbf{0.99} &  \textbf{0.99}	&  \textbf{0.99} \\
		&LR (Ours)  & \textbf{0.99} &  \textbf{0.99} &  \textbf{0.99}	&  \textbf{0.99} &  \textbf{0.99} &  \textbf{0.99}	&  \textbf{0.99}\\
        \hline

  \end{tabular}

  \end{adjustbox}
    \caption{Comparison between state-of-the-art defense methods in terms of detection AUROC on ImageNet.}
    \label{tab:table2}
\end{table}


We conducted extensive comparisons with the existing computationally-efficient defenses. 
In Table \ref{tab:main-results} we report the AUROC scores for JPEG, Random, Deflect, Denoise, WDSR, FS and our method against BIM, PGD, PIF, APGD, ANDA, VMI and VNI attacks targeting 7 models, namely VGG19, ResNet50, InceptionV3, ViT, DeiT, LeViT, on two datasets, ImageNet and CIFAR-100.
In every experimental setting, with different attacks, target models, and datasets, our proposed method outperforms these defense methods by a wide margin. Compared to LR's average AUROC score of $0.98$, the best performance among the existing efficient methods remains at $0.64$. 
More importantly, our method is robust {across varying target models and attack types}, as indicated by its small standard deviation.  While the other defense methods are effective against certain target-attack combinations, they fail to generalize this to a wide range of settings. For instance, on ImageNet, JPEG, Random, and FS perform better with transformer models, however they rarely exceed the random guess performance with the CNN models.

Due to the high computational requirements of VLAD and EPS-AD, we benchmark these methods in a smaller setting, where 1,000 images and 3 target models are used with the same 7 attacks and the results are given in Table \ref{tab:table2}. In all of the test cases, VLAD has an average AUROC score of $0.88$. It performs worst against the attacks targeted ResNet50 where its performance varies between $0.82$ and $0.77$. On the other hand, similarly to the results demonstrated in Table \ref{tab:main-results}, LR proves its robustness to different attack and target combinations with the average AUROC score of $0.99$. EPS-AD has a similar performance to our method where it successfully detects attacks with an average AUROC score of $0.99$. However, compared to LR, both VLAD and EPS-AD have considerably large computational cost, which limits their real-world usage. In the next section, we discuss the computational efficienty and real-world applicability
of all defense methods considered in this section.

\subsection{Computational Efficiency}
\label{sec:comp-cost}

\begin{figure}[t]
    \centering
    \includegraphics[width=0.55\linewidth]{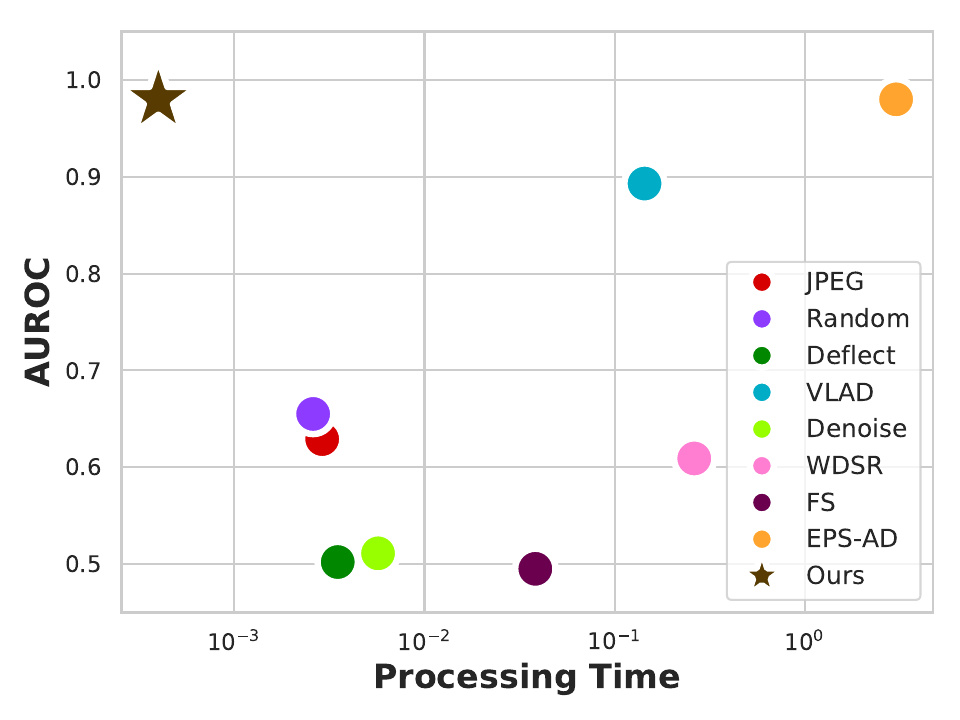}
    \vspace{-2mm}
  \caption{The proposed detector is both universal (high AUROC across all scenarios) and lightweight (fastest among all existing defenses).}%
    \label{fig:pt_avg}
    \vspace{-3mm}
\end{figure}



Real-time attack detection is a crucial aspect of many real-world systems. A detection mechanism must {operate consistently} alongside the DNN model to ensure timely identification of adversarial examples. Consequently, an ideal detector should be computationally efficient. In this section, we compare the computational costs of the defense methods evaluated in our experiments. For each defense method, we processed 1,000 samples from ImageNet and calculated the average processing time per sample. Figure \ref{fig:pt_avg} presents the processing time per sample (PTS) in seconds vs. AUROC for each defense method. Our method is the fastest, with a processing time of just 0.0004 seconds. In contrast, VLAD, WSDR, and EPS-AD are the slowest, with PTS values of 0.1431, 0.2611, and 2.9998 seconds, respectively. Our method demonstrates exceptional detection performance {and is also the fastest}. Notably, our method runs $\mathbf{1.3 \times 10^6}$ times faster than EPS-AD, which achieves similar detection performance to ours. The processing times are measured using a desktop computer with an NVIDIA 4090 GPU, AMD Ryzen 9 7950X CPU, and 64 GB RAM.


\subsection{Performance under Varying Attack Strength}

\begin{figure}[t]
    \centering
    \includegraphics[width=0.55\linewidth]{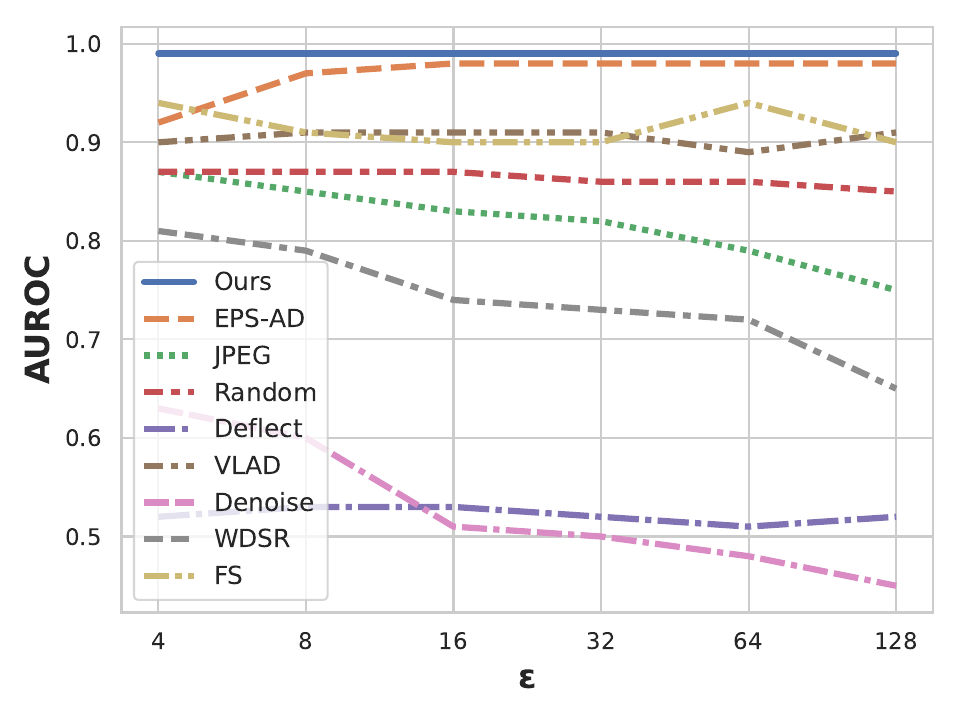}
    \vspace{-3mm}
  \caption{Detection performance vs attack strength.}
    \label{fig:attack-str}
\end{figure}

Adversarial attacks often use a parameter $\epsilon$ to adjust the amount of perturbation on adversarial examples. For the experiments in Section \ref{sec:exps}, we used the default attack settings that are proposed by the authors. In this study, we investigate the performance of defense methods against PGD attack with different $\epsilon$ values. In the original implementation of PGD{,} attack strength parameter $\epsilon$ is set to 8. In addition to the original $\epsilon$, we also generated attacks by setting $\epsilon$ to 4, 16, 32, 64 and 128.
Figure \ref{fig:attack-str} plots the performances of defense methods for ViT-PGD model-attack pair. Our method performs the same for every $\epsilon$ value, since LR depends on the {perturbation’s effects} on DNN layers, {a} significant performance drop is not expected due to the changes in attack strength. EPS-AD performs lower on weaker attacks, especially when $\epsilon$ is set to 4, their performance drops considerably. In remaining defense methods, while VLAD and Deflect experience only small changes under stronger attacks, {whereas the others} are greatly affected by the $\epsilon$ value. 



\subsection{Performance against Adaptive Attacks}
\label{sec:aa}

\begin{table}
  \centering
  \begin{tabular}{ c | c  c  c  c  c  c }
    Model & VGG19 & ResNet & IncV3 & ViT & DeiT & LeViT \\
    \hline
    AUROC & 0.99 & 0.97 & 0.96 & 0.99 & 0.99  & 0.98  \\

  \end{tabular}
  \caption{LR's performance against AutoAttack.}
  \label{tab:autoattack}
  
\end{table}

AutoAttack is an {ensemble-based method combining multiple attacks}. In recent works, it is used as an adaptive atttack \citep{eps-ad} to evaluate detectors. In Table \ref{tab:autoattack}, we provide LR's performance against AutoAttack targetting 6 models. Similarly to the results in Section \ref{sec:white}, LR performs with an average of 0.98\% AUROC score. 

Since AutoAttack does not use any knowledge of the detector, we design a PGD-based targeted adaptive attack, following a similar approach to \citep{contra}, with the objective function
\begin{equation}
    \max L_{Classifier} - \lambda \cdot L_{LR},
\end{equation}
where $L_{Classifier}$ and $L_{LR}$ denote the loss functions of the target model and LR detector, respectively.
The sign of LR’s loss function is (-) since LR identifies an input as clean when the loss is low. 
Note that, since this threat model has the complete knowledge of the detector, it is extremely challenging and destructive for the detector. To cope with this challenge, we propose randomly concatenating the layer vectors described in Equation \eqref{eq:v_vector} at inference time. 

In Table \ref{tab:adaptiveattack}, with ViT as the target model and the experimental settings proposed in \citep{contra} where number of iterations are set as 200 and $\lambda$ is set as 1, we report LR's performance under adaptive PGD attack with varying attack budgets $\epsilon \in [4, 8, 16, 32, 64, 128]$. {LR achieves an average AUROC score of 80\%, even under this challenging adaptive threat model}


\begin{table}[t]
  \centering
  \begin{tabular}{ c | c  c  c  c  c  c }
    $\epsilon$ & 4 & 8 & 16 & 32 & 64 & 128 \\
    \hline
    AUROC & 0.81 & 0.78 & 0.79 & 0.80 & 0.81  &  0.81 \\

  \end{tabular}
  \caption{LR's performance under varying strengts of PGD-based adaptive attack where $\lambda$ is set to 1.}
  \label{tab:adaptiveattack}
  \vspace{-2mm}
\end{table}

\section{Applicability in Other Domains}

LR is a universal detector applicable in various domains where DNNs are used. Here, we demonstrate its performance in two {additional} domains, video action recognition and speech recognition. In Supplementary \ref{ap:more_work} and \ref{ap:details_ab}, we provide more information about the experimental details used in this section. We {also} present additional results on another application, traffic sign recognition, in Supplementary \ref{ap:traffic}.

\subsection{LR for Video Action Recognition}
\label{sec:ab1}

In this section, we implement LR against video action recognition attacks and compare its performance {to} existing defenses designed for action recognition models, namely Advit \citep{advit}, Shuffle \citep{shuffle}, and  VLAD \citep{vlad}. In the experiments, PGD-v attack \citep{vlad} and Flick attack \citep{flick} are used to target MVIT \citep{mvit} and X3D \citep{x3d}. The experimental settings in \cite{vlad} on Kinetics-400 \citep{k400} dataset are followed. The results in Table \ref{tab:vid_results} show that LR outperforms the other defenses with an average {AUROC} of 0.93\%, followed by VLAD which achieves 0.91\%.

\begin{table}[t]
  \centering
  
  \begin{tabular}{c  c | c  c  c  c }
    
    &  & Advit & Shuffle & VLAD & LR (ours)  \\
    \hline

            \multirow{2}{*}{MVIT}
            &PGD-v  & 0.93 & 0.98	& 0.93	&  \textbf{0.99}\\
		&Flick  &0.34 & 0.65 & 0.87	&  \textbf{0.89}\\
		
  \hline 

            \multirow{2}{*}{X3D}
            & PGD-v   & 0.92 & 0.76 	&  \textbf{0.97}	& 0.95 \\
		& Flick \ &0.54 &  0.59 &	0.90	&  \textbf{0.92}\\
		
        \hline
            & Average & 0.68 &	0.74 &	0.91 &	 \textbf{0.93}\\

  \end{tabular}
  \caption{AUROC scores against attack methods targeting video action recognition models.}
  \label{tab:vid_results}
\end{table}

\subsection{LR for Speech Recognition}
\label{sec:ab2}

\begin{figure}[t]
    \centering
    \includegraphics[width=.8\columnwidth]{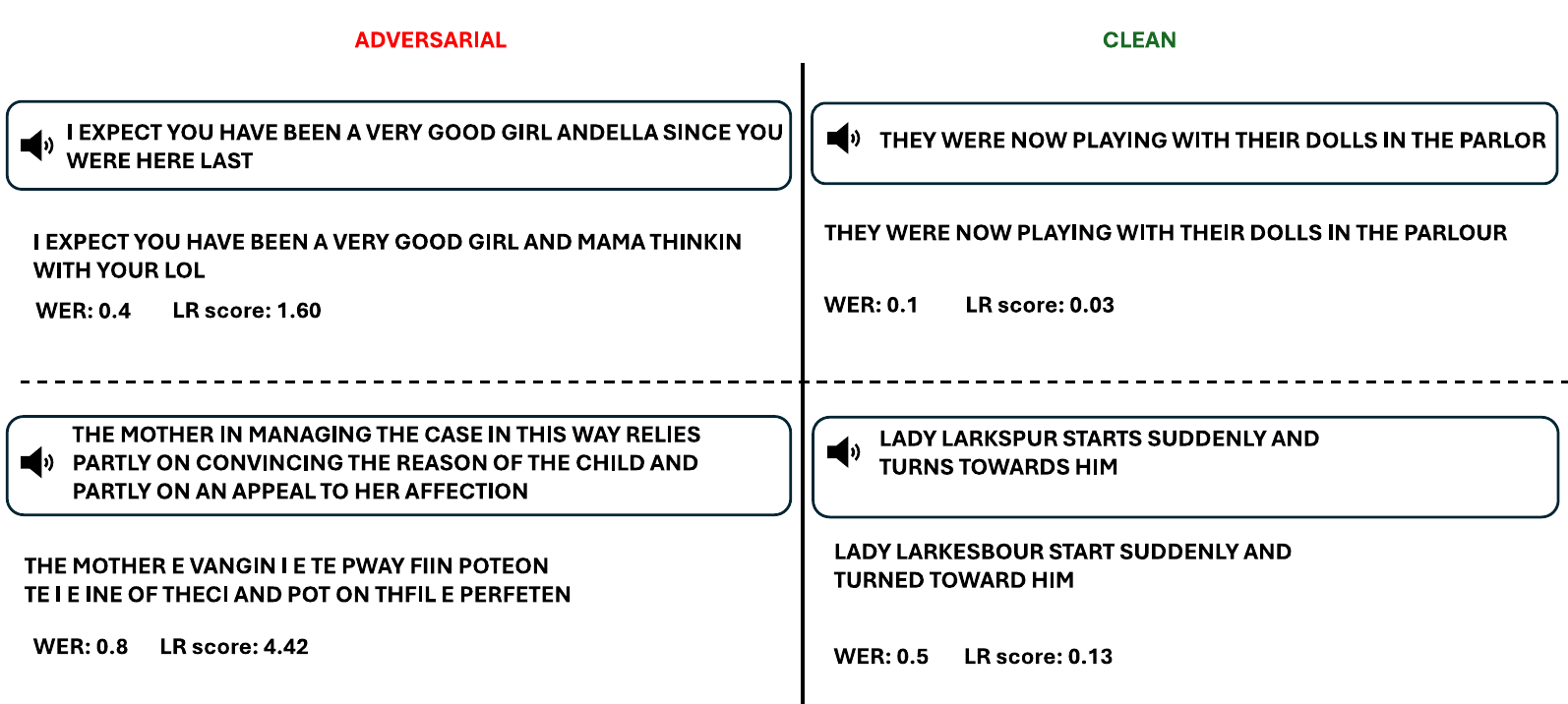}
    \caption{Clean and adversarial samples for speech recognition with ground truth in box, recognized text, word error rate (WER), and MSE of proposed LR.}
    \label{fig:waw}
\end{figure}

To demonstrate LR's wide applicability {beyond} computer vision, we also {evaluate} its performance against a speech recognition attack. As target model, we use Wav2vec \citep{w2v} model, which is trained on LibriSpeech \citep{libri} dataset. Model is attacked with FGSM \citep{fgsm}. LR achieves an average AUROC of $0.99$. Figure \ref{fig:waw} illustrates some attacked and clean samples from LibriSpeech, along with the MSE values of LR. As shown in the figure, LR can even detect stealthy attacks which {cause minimal increase in} word error rate (WER) while distorting the recognized speech. 
Remarkably, while the WER of the first adversarial example is lower than {that} of the second clean example ($0.4$ vs. $0.5$), the LR loss for this stealthy adversarial sample is more than ten folds greater than that of the second clean example ($1.6$ vs $0.13$). 


\section{Conclusion}

Although there are effective defense methods for specific model-attack combinations, their success {does} not generalize to all popular models and attacks. In this work, we filled this gap by introducing Layer Regression (LR), the first universal method for detecting adversarial examples. In addition to its universality, LR is much faster than the existing defense methods. By analyzing the common objectives of attacks and the sequential layer-based nature of DNNs, we showed that {adversarial samples have greater impact on the final layer than on the initial layers.} Motivated by this, LR trains a lightweight multi-layer perceptron (MLP) on clean samples to estimate the feature vector using a combination of outputs from early layers. The estimation error of LR for adversarial samples is typically much higher than the error for clean samples. {Through} extensive experiments, we showed that LR outperforms the existing defenses in image recognition by a wide margin in terms of speed and detection performance, and also {delivers} highly effective {results} in distinct domains, namely video action recognition and speech recognition. While inference time randomization of input order in LR is shown to be a promising technique {for addressing adaptive attacks that are} fully aware of LR, {further research is needed to explore} its full potential in this challenging scenario.






\bibliography{tmlr}

\begin{thebibliography}{59}
\providecommand{\natexlab}[1]{#1}
\providecommand{\url}[1]{\texttt{#1}}
\expandafter\ifx\csname urlstyle\endcsname\relax
  \providecommand{\doi}[1]{doi: #1}\else
  \providecommand{\doi}{doi: \begingroup \urlstyle{rm}\Url}\fi

\bibitem[Aydemir et~al.(2018)Aydemir, Temizel, and Temizel]{jpeg2}
Ayse~Elvan Aydemir, Alptekin Temizel, and Tugba~Taskaya Temizel.
\newblock The effects of jpeg and jpeg2000 compression on attacks using adversarial examples.
\newblock \emph{arXiv preprint arXiv:1803.10418}, 2018.

\bibitem[Bai et~al.(2021)Bai, Luo, Zhao, Wen, and Wang]{adv_training}
Tao Bai, Jinqi Luo, Jun Zhao, Bihan Wen, and Qian Wang.
\newblock Recent advances in adversarial training for adversarial robustness.
\newblock \emph{arXiv preprint arXiv:2102.01356}, 2021.

\bibitem[Chen et~al.(2017)Chen, Zhang, Sharma, Yi, and Hsieh]{chen2017zoo}
Pin-Yu Chen, Huan Zhang, Yash Sharma, Jinfeng Yi, and Cho-Jui Hsieh.
\newblock Zoo: Zeroth order optimization based black-box attacks to deep neural networks without training substitute models.
\newblock In \emph{Proceedings of the 10th ACM workshop on artificial intelligence and security}, pp.\  15--26, 2017.

\bibitem[Croce \& Hein(2020{\natexlab{a}})Croce and Hein]{aa}
Francesco Croce and Matthias Hein.
\newblock Reliable evaluation of adversarial robustness with an ensemble of diverse parameter-free attacks.
\newblock In \emph{International conference on machine learning}, pp.\  2206--2216. PMLR, 2020{\natexlab{a}}.

\bibitem[Croce \& Hein(2020{\natexlab{b}})Croce and Hein]{apgd}
Francesco Croce and Matthias Hein.
\newblock Reliable evaluation of adversarial robustness with an ensemble of diverse parameter-free attacks.
\newblock In \emph{International conference on machine learning}, pp.\  2206--2216. PMLR, 2020{\natexlab{b}}.

\bibitem[Cucu et~al.(2023)Cucu, Valenzise, St{\u{a}}nescu, Ghergulescu, G{\u{a}}in{\u{a}}, and Gu{\c{s}}i{\c{t}}{\u{a}}]{jpeg1}
Adelina-Valentina Cucu, Giuseppe Valenzise, Daniela St{\u{a}}nescu, Ioana Ghergulescu, Lucian~Ionel G{\u{a}}in{\u{a}}, and Bianca Gu{\c{s}}i{\c{t}}{\u{a}}.
\newblock Defense method against adversarial attacks using jpeg compression and one-pixel attack for improved dataset security.
\newblock In \emph{2023 27th International Conference on System Theory, Control and Computing (ICSTCC)}, pp.\  523--527. IEEE, 2023.

\bibitem[Das et~al.(2018)Das, Shanbhogue, Chen, Hohman, Li, Chen, Kounavis, and Chau]{jpeg3}
Nilaksh Das, Madhuri Shanbhogue, Shang-Tse Chen, Fred Hohman, Siwei Li, Li~Chen, Michael~E Kounavis, and Duen~Horng Chau.
\newblock Shield: Fast, practical defense and vaccination for deep learning using jpeg compression.
\newblock In \emph{Proceedings of the 24th ACM SIGKDD International Conference on Knowledge Discovery \& Data Mining}, pp.\  196--204, 2018.

\bibitem[Dosovitskiy(2020)]{vit}
Alexey Dosovitskiy.
\newblock An image is worth 16x16 words: Transformers for image recognition at scale.
\newblock \emph{arXiv preprint arXiv:2010.11929}, 2020.

\bibitem[Fan et~al.(2021)Fan, Xiong, Mangalam, Li, Yan, Malik, and Feichtenhofer]{mvit}
Haoqi Fan, Bo~Xiong, Karttikeya Mangalam, Yanghao Li, Zhicheng Yan, Jitendra Malik, and Christoph Feichtenhofer.
\newblock Multiscale vision transformers.
\newblock In \emph{Proceedings of the IEEE/CVF international conference on computer vision}, pp.\  6824--6835, 2021.

\bibitem[Fang et~al.(2024)Fang, Wang, Huang, and Jing]{anda}
Zhengwei Fang, Rui Wang, Tao Huang, and Liping Jing.
\newblock Strong transferable adversarial attacks via ensembled asymptotically normal distribution learning.
\newblock In \emph{Proceedings of the IEEE/CVF Conference on Computer Vision and Pattern Recognition}, pp.\  24841--24850, 2024.

\bibitem[Feichtenhofer(2020)]{x3d}
Christoph Feichtenhofer.
\newblock X3d: Expanding architectures for efficient video recognition.
\newblock In \emph{Proceedings of the IEEE/CVF conference on computer vision and pattern recognition}, pp.\  203--213, 2020.

\bibitem[Gao et~al.(2020)Gao, Zhang, Song, Liu, and Shen]{pif}
Lianli Gao, Qilong Zhang, Jingkuan Song, Xianglong Liu, and Heng~Tao Shen.
\newblock Patch-wise attack for fooling deep neural network.
\newblock In \emph{Computer Vision--ECCV 2020: 16th European Conference, Glasgow, UK, August 23--28, 2020, Proceedings, Part XXVIII 16}, pp.\  307--322. Springer, 2020.

\bibitem[Goodfellow et~al.(2014)Goodfellow, Shlens, and Szegedy]{fgsm}
Ian~J Goodfellow, Jonathon Shlens, and Christian Szegedy.
\newblock Explaining and harnessing adversarial examples.
\newblock \emph{arXiv preprint arXiv:1412.6572}, 2014.

\bibitem[Graham et~al.(2021)Graham, El-Nouby, Touvron, Stock, Joulin, J{\'e}gou, and Douze]{levit}
Benjamin Graham, Alaaeldin El-Nouby, Hugo Touvron, Pierre Stock, Armand Joulin, Herv{\'e} J{\'e}gou, and Matthijs Douze.
\newblock Levit: a vision transformer in convnet's clothing for faster inference.
\newblock In \emph{Proceedings of the IEEE/CVF international conference on computer vision}, pp.\  12259--12269, 2021.

\bibitem[Grosse et~al.(2017)Grosse, Manoharan, Papernot, Backes, and McDaniel]{grosse2017statistical}
Kathrin Grosse, Praveen Manoharan, Nicolas Papernot, Michael Backes, and Patrick McDaniel.
\newblock On the (statistical) detection of adversarial examples.
\newblock \emph{arXiv preprint arXiv:1702.06280}, 2017.

\bibitem[He et~al.(2015)He, Zhang, Ren, and Sun]{resnet}
Kaiming He, Xiangyu Zhang, Shaoqing Ren, and Jian Sun.
\newblock Deep residual learning for image recognition. arxiv e-prints.
\newblock \emph{arXiv preprint arXiv:1512.03385}, 10, 2015.

\bibitem[Hsiao et~al.(2024)Hsiao, Huang, Ni, Lin, Shuai, Li, and Cheng]{light}
Teng-Fang Hsiao, Bo-Lun Huang, Zi-Xiang Ni, Yan-Ting Lin, Hong-Han Shuai, Yung-Hui Li, and Wen-Huang Cheng.
\newblock Natural light can also be dangerous: Traffic sign misinterpretation under adversarial natural light attacks.
\newblock In \emph{Proceedings of the IEEE/CVF Winter Conference on Applications of Computer Vision}, pp.\  3915--3924, 2024.

\bibitem[Hwang et~al.(2023)Hwang, Zhang, Choi, Hsieh, and Lee]{shuffle}
Jaehui Hwang, Huan Zhang, Jun-Ho Choi, Cho-Jui Hsieh, and Jong-Seok Lee.
\newblock Temporal shuffling for defending deep action recognition models against adversarial attacks.
\newblock \emph{Neural Networks}, 2023.

\bibitem[Ilyas et~al.(2018)Ilyas, Engstrom, Athalye, and Lin]{ilyas2018black}
Andrew Ilyas, Logan Engstrom, Anish Athalye, and Jessy Lin.
\newblock Black-box adversarial attacks with limited queries and information.
\newblock In \emph{International Conference on Machine Learning}, pp.\  2137--2146. PMLR, 2018.

\bibitem[Kay et~al.(2017)Kay, Carreira, Simonyan, Zhang, Hillier, Vijayanarasimhan, Viola, Green, Back, Natsev, et~al.]{k400}
Will Kay, Joao Carreira, Karen Simonyan, Brian Zhang, Chloe Hillier, Sudheendra Vijayanarasimhan, Fabio Viola, Tim Green, Trevor Back, Paul Natsev, et~al.
\newblock The kinetics human action video dataset.
\newblock \emph{arXiv preprint arXiv:1705.06950}, 2017.

\bibitem[Krizhevsky et~al.(2009)Krizhevsky, Hinton, et~al.]{cifar}
Alex Krizhevsky, Geoffrey Hinton, et~al.
\newblock Learning multiple layers of features from tiny images.
\newblock 2009.

\bibitem[Kurakin et~al.(2018)Kurakin, Goodfellow, and Bengio]{bim}
Alexey Kurakin, Ian~J Goodfellow, and Samy Bengio.
\newblock Adversarial examples in the physical world.
\newblock In \emph{Artificial intelligence safety and security}, pp.\  99--112. Chapman and Hall/CRC, 2018.

\bibitem[Liao et~al.(2018)Liao, Liang, Dong, Pang, Hu, and Zhu]{denoise1}
Fangzhou Liao, Ming Liang, Yinpeng Dong, Tianyu Pang, Xiaolin Hu, and Jun Zhu.
\newblock Defense against adversarial attacks using high-level representation guided denoiser.
\newblock In \emph{Proceedings of the IEEE conference on computer vision and pattern recognition}, pp.\  1778--1787, 2018.

\bibitem[Madry et~al.(2017)Madry, Makelov, Schmidt, Tsipras, and Vladu]{pgd}
Aleksander Madry, Aleksandar Makelov, Ludwig Schmidt, Dimitris Tsipras, and Adrian Vladu.
\newblock Towards deep learning models resistant to adversarial attacks.
\newblock \emph{arXiv preprint arXiv:1706.06083}, 2017.

\bibitem[Metzen et~al.(2017)Metzen, Genewein, Fischer, and Bischoff]{metzen2017detecting}
Jan~Hendrik Metzen, Tim Genewein, Volker Fischer, and Bastian Bischoff.
\newblock On detecting adversarial perturbations.
\newblock \emph{arXiv preprint arXiv:1702.04267}, 2017.

\bibitem[Mumcu \& Yilmaz(2024{\natexlab{a}})Mumcu and Yilmaz]{mumcu2024sequential}
Furkan Mumcu and Yasin Yilmaz.
\newblock Sequential architecture-agnostic black-box attack design and analysis.
\newblock \emph{Pattern Recognition}, 147:\penalty0 110066, 2024{\natexlab{a}}.

\bibitem[Mumcu \& Yilmaz(2024{\natexlab{b}})Mumcu and Yilmaz]{vilas}
Furkan Mumcu and Yasin Yilmaz.
\newblock Fast and lightweight vision-language model for adversarial traffic sign detection.
\newblock \emph{Electronics}, 13\penalty0 (11):\penalty0 2172, 2024{\natexlab{b}}.

\bibitem[Mumcu \& Yilmaz(2024{\natexlab{c}})Mumcu and Yilmaz]{vlad}
Furkan Mumcu and Yasin Yilmaz.
\newblock Multimodal attack detection for action recognition models.
\newblock In \emph{Proceedings of the IEEE/CVF Conference on Computer Vision and Pattern Recognition}, pp.\  2967--2976, 2024{\natexlab{c}}.

\bibitem[Mumcu et~al.(2022)Mumcu, Doshi, and Yilmaz]{mumcu2022adversarial}
Furkan Mumcu, Keval Doshi, and Yasin Yilmaz.
\newblock Adversarial machine learning attacks against video anomaly detection systems.
\newblock In \emph{Proceedings of the IEEE/CVF Conference on Computer Vision and Pattern Recognition}, pp.\  206--213, 2022.

\bibitem[Mustafa et~al.(2019)Mustafa, Khan, Hayat, Shen, and Shao]{wdsr}
Aamir Mustafa, Salman~H Khan, Munawar Hayat, Jianbing Shen, and Ling Shao.
\newblock Image super-resolution as a defense against adversarial attacks.
\newblock \emph{IEEE Transactions on Image Processing}, 29:\penalty0 1711--1724, 2019.

\bibitem[Panayotov et~al.(2015)Panayotov, Chen, Povey, and Khudanpur]{libri}
Vassil Panayotov, Guoguo Chen, Daniel Povey, and Sanjeev Khudanpur.
\newblock Librispeech: an asr corpus based on public domain audio books.
\newblock In \emph{2015 IEEE international conference on acoustics, speech and signal processing (ICASSP)}, pp.\  5206--5210. IEEE, 2015.

\bibitem[Pang et~al.(2018)Pang, Xu, Du, Chen, and Zhu]{pang2018rce}
Tianyu Pang, Kun Xu, Chao Du, Ning Chen, and Jun Zhu.
\newblock Towards robust detection of adversarial examples.
\newblock In \emph{NeurIPS}, 2018.

\bibitem[Papernot \& McDaniel(2018)Papernot and McDaniel]{papernot2018deep}
Nicolas Papernot and Patrick McDaniel.
\newblock Deep k-nearest neighbors: Towards confident, interpretable and robust deep learning.
\newblock \emph{arXiv preprint arXiv:1803.04765}, 2018.

\bibitem[Papernot et~al.(2017)Papernot, McDaniel, Goodfellow, Jha, Celik, and Swami]{papernot2017practical}
Nicolas Papernot, Patrick McDaniel, Ian Goodfellow, Somesh Jha, Z~Berkay Celik, and Ananthram Swami.
\newblock Practical black-box attacks against machine learning.
\newblock In \emph{Proceedings of the 2017 ACM on Asia conference on computer and communications security}, pp.\  506--519, 2017.

\bibitem[Paszke et~al.(2017)Paszke, Gross, Chintala, Chanan, Yang, DeVito, Lin, Desmaison, Antiga, and Lerer]{paszke2017automatic}
Adam Paszke, Sam Gross, Soumith Chintala, Gregory Chanan, Edward Yang, Zachary DeVito, Zeming Lin, Alban Desmaison, Luca Antiga, and Adam Lerer.
\newblock Automatic differentiation in pytorch.
\newblock 2017.

\bibitem[Pony et~al.(2021)Pony, Naeh, and Mannor]{flick}
Roi Pony, Itay Naeh, and Shie Mannor.
\newblock Over-the-air adversarial flickering attacks against video recognition networks.
\newblock In \emph{Proceedings of the IEEE/CVF Conference on Computer Vision and Pattern Recognition}, pp.\  515--524, 2021.

\bibitem[Prakash et~al.(2018)Prakash, Moran, Garber, DiLillo, and Storer]{deflection}
Aaditya Prakash, Nick Moran, Solomon Garber, Antonella DiLillo, and James Storer.
\newblock Deflecting adversarial attacks with pixel deflection.
\newblock In \emph{Proceedings of the IEEE conference on computer vision and pattern recognition}, pp.\  8571--8580, 2018.

\bibitem[Radford et~al.(2021)Radford, Kim, Hallacy, Ramesh, Goh, Agarwal, Sastry, Askell, Mishkin, Clark, et~al.]{clip}
Alec Radford, Jong~Wook Kim, Chris Hallacy, Aditya Ramesh, Gabriel Goh, Sandhini Agarwal, Girish Sastry, Amanda Askell, Pamela Mishkin, Jack Clark, et~al.
\newblock Learning transferable visual models from natural language supervision.
\newblock In \emph{International conference on machine learning}, pp.\  8748--8763. PMLR, 2021.

\bibitem[Russakovsky et~al.(2015)Russakovsky, Deng, Su, Krause, Satheesh, Ma, Huang, Karpathy, Khosla, Bernstein, Berg, and Fei-Fei]{imagenet}
Olga Russakovsky, Jia Deng, Hao Su, Jonathan Krause, Sanjeev Satheesh, Sean Ma, Zhiheng Huang, Andrej Karpathy, Aditya Khosla, Michael Bernstein, Alexander~C. Berg, and Li~Fei-Fei.
\newblock {ImageNet Large Scale Visual Recognition Challenge}.
\newblock \emph{International Journal of Computer Vision (IJCV)}, 115\penalty0 (3):\penalty0 211--252, 2015.
\newblock \doi{10.1007/s11263-015-0816-y}.

\bibitem[Salman et~al.(2020)Salman, Sun, Yang, Kapoor, and Kolter]{denoise3}
Hadi Salman, Mingjie Sun, Greg Yang, Ashish Kapoor, and J~Zico Kolter.
\newblock Denoised smoothing: A provable defense for pretrained classifiers.
\newblock \emph{Advances in Neural Information Processing Systems}, 33:\penalty0 21945--21957, 2020.

\bibitem[Schneider et~al.(2019)Schneider, Baevski, Collobert, and Auli]{w2v}
Steffen Schneider, Alexei Baevski, Ronan Collobert, and Michael Auli.
\newblock wav2vec: Unsupervised pre-training for speech recognition.
\newblock \emph{arXiv preprint arXiv:1904.05862}, 2019.

\bibitem[Simonyan \& Zisserman(2014)Simonyan and Zisserman]{vgg}
Karen Simonyan and Andrew Zisserman.
\newblock Very deep convolutional networks for large-scale image recognition.
\newblock \emph{arXiv preprint arXiv:1409.1556}, 2014.

\bibitem[Stallkamp et~al.(2012)Stallkamp, Schlipsing, Salmen, and Igel]{gtsrb}
J.~Stallkamp, M.~Schlipsing, J.~Salmen, and C.~Igel.
\newblock Man vs. computer: Benchmarking machine learning algorithms for traffic sign recognition.
\newblock \emph{Neural Networks}, \penalty0 (0):\penalty0 --, 2012.
\newblock ISSN 0893-6080.
\newblock \doi{10.1016/j.neunet.2012.02.016}.
\newblock URL \url{http://www.sciencedirect.com/science/article/pii/S0893608012000457}.

\bibitem[Szegedy et~al.(2015)Szegedy, Liu, Jia, Sermanet, Reed, Anguelov, Erhan, Vanhoucke, and Rabinovich]{inception}
Christian Szegedy, Wei Liu, Yangqing Jia, Pierre Sermanet, Scott Reed, Dragomir Anguelov, Dumitru Erhan, Vincent Vanhoucke, and Andrew Rabinovich.
\newblock Going deeper with convolutions.
\newblock In \emph{Proceedings of the IEEE conference on computer vision and pattern recognition}, pp.\  1--9, 2015.

\bibitem[Tian et~al.(2018)Tian, Yang, and Cai]{tian2018detecting}
Yu~Tian, Xiaopeng Yang, and Yuanqing Cai.
\newblock Detecting adversarial examples through image transformations.
\newblock In \emph{AAAI}, 2018.

\bibitem[Touvron et~al.(2021)Touvron, Cord, Douze, Massa, Sablayrolles, and J{\'e}gou]{deit}
Hugo Touvron, Matthieu Cord, Matthijs Douze, Francisco Massa, Alexandre Sablayrolles, and Herv{\'e} J{\'e}gou.
\newblock Training data-efficient image transformers \& distillation through attention.
\newblock In \emph{International conference on machine learning}, pp.\  10347--10357. PMLR, 2021.

\bibitem[Tran et~al.(2019)Tran, Wang, Torresani, and Feiszli]{csn}
Du~Tran, Heng Wang, Lorenzo Torresani, and Matt Feiszli.
\newblock Video classification with channel-separated convolutional networks.
\newblock In \emph{Proceedings of the IEEE/CVF international conference on computer vision}, pp.\  5552--5561, 2019.

\bibitem[Wang \& He(2021)Wang and He]{vmi_vni}
Xiaosen Wang and Kun He.
\newblock Enhancing the transferability of adversarial attacks through variance tuning.
\newblock In \emph{Proceedings of the IEEE/CVF conference on computer vision and pattern recognition}, pp.\  1924--1933, 2021.

\bibitem[Wightman(2019)]{rw2019timm}
Ross Wightman.
\newblock Pytorch image models.
\newblock \url{https://github.com/rwightman/pytorch-image-models}, 2019.

\bibitem[Xiao et~al.(2019)Xiao, Deng, Li, Lee, Edwards, Yi, Song, Liu, and Molloy]{advit}
Chaowei Xiao, Ruizhi Deng, Bo~Li, Taesung Lee, Benjamin Edwards, Jinfeng Yi, Dawn Song, Mingyan Liu, and Ian Molloy.
\newblock Advit: Adversarial frames identifier based on temporal consistency in videos.
\newblock In \emph{Proceedings of the IEEE/CVF International Conference on Computer Vision}, pp.\  3968--3977, 2019.

\bibitem[Xie et~al.(2017)Xie, Wang, Zhang, Ren, and Yuille]{random}
Cihang Xie, Jianyu Wang, Zhishuai Zhang, Zhou Ren, and Alan Yuille.
\newblock Mitigating adversarial effects through randomization.
\newblock \emph{arXiv preprint arXiv:1711.01991}, 2017.

\bibitem[Xiong et~al.(2022)Xiong, Eappen, Zhu, and Jagannathan]{denoise2}
Zikang Xiong, Joe Eappen, He~Zhu, and Suresh Jagannathan.
\newblock Defending observation attacks in deep reinforcement learning via detection and denoising.
\newblock In \emph{Joint European Conference on Machine Learning and Knowledge Discovery in Databases}, pp.\  235--250. Springer, 2022.

\bibitem[Xu(2017{\natexlab{a}})]{fs}
W~Xu.
\newblock Feature squeezing: Detecting adversarial exa mples in deep neural networks.
\newblock \emph{arXiv preprint arXiv:1704.01155}, 2017{\natexlab{a}}.

\bibitem[Xu(2017{\natexlab{b}})]{squeezing}
W~Xu.
\newblock Feature squeezing: Detecting adversarial exa mples in deep neural networks.
\newblock \emph{arXiv preprint arXiv:1704.01155}, 2017{\natexlab{b}}.

\bibitem[Yang et~al.(2022)Yang, Gao, Li, Lai, and Xu]{contra}
Yijun Yang, Ruiyuan Gao, Yu~Li, Qiuxia Lai, and Qiang Xu.
\newblock What you see is not what the network infers: Detecting adversarial examples based on semantic contradiction.
\newblock \emph{arXiv preprint arXiv:2201.09650}, 2022.

\bibitem[Ye et~al.(2021)Ye, Yin, Yan, and Ge]{tpatch}
Bin Ye, Huilin Yin, Jun Yan, and Wanchen Ge.
\newblock Patch-based attack on traffic sign recognition.
\newblock In \emph{2021 IEEE International Intelligent Transportation Systems Conference (ITSC)}, pp.\  164--171. IEEE, 2021.

\bibitem[{\.Z}elasko et~al.(2021){\.Z}elasko, Joshi, Shao, Villalba, Trmal, Dehak, and Khudanpur]{zelasko2021adversarial}
Piotr {\.Z}elasko, Sonal Joshi, Yiwen Shao, Jesus Villalba, Jan Trmal, Najim Dehak, and Sanjeev Khudanpur.
\newblock Adversarial attacks and defenses for speech recognition systems.
\newblock \emph{arXiv preprint arXiv:2103.17122}, 2021.

\bibitem[Zhang et~al.(2023)Zhang, Liu, Yang, Yang, Li, Han, and Tan]{eps-ad}
Shuhai Zhang, Feng Liu, Jiahao Yang, Yifan Yang, Changsheng Li, Bo~Han, and Mingkui Tan.
\newblock Detecting adversarial data by probing multiple perturbations using expected perturbation score.
\newblock In \emph{International conference on machine learning}, pp.\  41429--41451. PMLR, 2023.

\bibitem[Zhao et~al.(2019)Zhao, Zhu, Liang, Shen, Zhang, and Chen]{zhao2019seeing}
Yue Zhao, Hong Zhu, Ruigang Liang, Qintao Shen, Shengzhi Zhang, and Kai Chen.
\newblock Seeing isn't believing: Towards more robust adversarial attack against real world object detectors.
\newblock In \emph{Proceedings of the 2019 ACM SIGSAC conference on computer and communications security}, pp.\  1989--2004, 2019.

\end{thebibliography}
\bibliographystyle{tmlr}

\clearpage
\appendix

\section{More Related Work from Other Domains}
\label{ap:more_work}

Adversarial examples are studied on several other domains including video action recognition \citep{flick}, traffic sign recognition \citep{light}, object detection \citep{zhao2019seeing}, video anomaly detection \citep{mumcu2022adversarial}, and speech recognition \citep{zelasko2021adversarial}. For instance, flickering attack (Flick) \citep{flick} tries to attack action recognition models by changing the RGB stream of the input videos. \cite{light} investigates the effects of natural light on traffic signs and use it to generate adversarial examples.

Advit \citep{advit} is a detection method introduced for videos. It generates pseudo frames using optical flow and evaluates the consistency between the outputs for original inputs and pseudo frames to detect attacks. Another defense method, Shuffle \citep{shuffle}, tries to increase the robustness of action recognition models by randomly shuffling the input frames. \citep{vilas} proposed to train a lightweight VLM and use it for adversarial traffic sign detection. In speech recognition, defenses like denoising or smoothing are studied against adversarial examples \citep{zelasko2021adversarial}. 
\section{LR's Performance Against $l_2$ and Targeted Attacks}
\label{ap:tab:targeted-l2}

\begin{table}[h]
  \centering
  
  \begin{tabular}{ c | c  c  c  c }
     & $PGD_{targeted}$ & $PGD_{l_2}$ & $APGD_{targeted}$ & $APGD_{l_2}$  \\

    \hline
    ResNet50  & 0.99 & 0.95  & 0.98 &  0.95  \\
    ViT  & 0.99 & 0.95  & 0.98 &  0.96 \\
  \end{tabular}
  \caption{LR's performance under $l_2$ and targeted attack settings. }
  \label{tab:targeted-l2}
  
\end{table}

We evaluate LR's performance under $l_2$ and targeted attack settings, specifically against $PGD_{targeted}$, $PGD_{l_2}$, $APGD_{targeted}$ and $APGD_{l_2}$ attacks using two different target models ResNet50 and ViT on ImageNet. The results are reported in Table \ref{tab:targeted-l2}. Similar to the white-box attack setting, LR performs an average AUROC score of 97\% which further demonstrates the success of our detection method in different types of attacks.

\section{More Details on Experimental Settings}

From the validation set of ImageNet \citep{imagenet}, we used 40,000 images
to train the detectors and the remaining 10,000 for testing. In CIFAR-100 \citep{cifar}, there are 100 classes, 500 training images and 100 testing images per class, resulting in a total of 50,000 training and
10,000 test images. During the tests, only the correctly classified images by the target models were used.  Total number for test images for each specific model and dataset is given in Table \ref{tab:totalimages}.

\begin{table}[h]
  \centering
  
  \begin{tabular}{ c | c  c  c  c  c  c}
     & VGG19 & ResNet50 & InceptionV3 & ViT & DeiT & LeViT\\

     
    \hline
    ImageNet  & 4257 &  7701 & 7410 & 8457  & 7907 & 7639 \\
    CIFAR-100  & 7045 & 8315  & 8089 & 8544 & 8667 & 8359 \\
  \end{tabular}
  \caption{Number of test images used for each model.}
  \label{tab:totalimages}
  
\end{table}

VLAD \citep{vlad} was originally proposed for video recognition attacks, with the initial implementation designed for videos consisting of 30 frames. In the experiments conducted in Section \ref{sec:exps}, we used the official VLAD implementation. However, instead of averaging the scores across 30 frames, we applied it to a single image.
\section{Effects of Layer Vector Choice} 
\label{ap:inpu_sel}

For the experiments in Section \ref{sec:exps}, we used three layer vectors to form $v$, i.e., $m=3$ in Eq. \ref{eq:v_vector}. Since the number of {possible} combinations for layer selection and slicing {creates an intractably large search space, we limited our optimization to a representative subset of options.} Table \ref{tab:layers} summarizes the results of a preliminary experiment on CIFAR-100, which is conducted with DeiT as the target model and PGD and PIF as the adversarial attacks. In this experiment, we considered 25 attention layers of the model and trained our detectors by forming $v$ in four different ways: (1) using only the first attention layer vector $a_1$, (2) using only the last attention layer vector $a_{25}$, (3) combining the vectors from fifth, sixth, and seventh attention vectors $[a_5, a_6, a_7]$, (4) and combining the vectors from eighth, thirteenth, and seventeenth attention vectors $[a_8, a_{13}, a_{17}]$. The best strategy turns out to combining vectors from early layers $[a_5, a_6, a_7]$ as it strikes a good balance in the trade-off between the two conflicting goals (Figure \ref{fig:v_sel}): proximity of $v(x)$ and $v(x^{adv})$, and accuracy and stability of estimator $f$. Additionally, having multiple layers and slicing functions can create randomness in each LR implementation, which makes developing an attack against our detection system harder.
The layer selection and slicing strategies that gave the best results reported in Table \ref{tab:main-results} are listed in Supplementary \ref{ap:subsets}. 

\begin{table}[h]
  \centering
  
  \begin{tabular}{ c | c  c  c  c }
    Attack & $a_1$ & $[a_5, a_6, a_7]$ & $[a_8, a_{13}, a_{17}]$ & $a_{n-2}$ \\
    \hline
    PGD & 0.697 & 0.867 & 0.751 & 0.567  \\
    PIF & 0.744  & 0.972 & 0.746 & 0.521 \\
  \end{tabular}
  \caption{Impact of layer selection on detection score.}
  \label{tab:layers}
  
\end{table}

\section{Layer Vector Selection Procedure}

According to our preliminary experiments, using vectors {exclusively } from {either} early or final layers negatively impacts performance. Therefore, we propose selecting three layer vectors randomly between 1/5 and 4/5 of all layers. The proposed procedure for selecting layer vectors is detailed in Algorithm \ref{alg:layers}.

\begin{algorithm}[h]
\caption{Layer Vector Selection Procedure}\label{alg:layers}
\begin{algorithmic}[1]

\Input{all layer vectors $v_n$,} 
\Output{selected layer vectors $v_m$.}

\State Get the layer size of all layer vectors $v_n$ :
\State $n  \gets len(v_n) $ 
 
\State Decide upper and lower bounds:
\State $a  \gets n/5 $ 
\State $b  \gets 4\cdot n/5 $ 

\State Slice the vectors between the upper and lower bounds:
\State $v_m  \gets v_n[a:b] $ 

\State Randomly pick 3 vectors from $v_m$:
\State $v_m \gets random.choices[v_m, 3]$

\State Return $v_m$

\end{algorithmic}
\end{algorithm}

\section{Selected Layer Vectors and Slicing Functions}
\label{ap:subsets}

A specific subset of layer vectors $a_r$, as described in Equation \eqref{eq:ar}, is chosen for each target model that is used during the experiments in Section \ref{sec:exps}. For the models, Pytorch Image Models (timm) \citep{rw2019timm} is used. 

We used the following layer subsets for the target models:
\begin{enumerate}
    \item Resnet50: Layers with the name \textit{conv2} were filtered, then among 15 \textit{conv2} layers, 5th, 8th and 13th layers, for all ImageNet and CIFAR-100 tests
    \item InceptionV3: Layers with the name \textit{conv} were filtered, then among 94 \textit{conv} layers, 15th, 25th and 35th layers, for all ImageNet and CIFAR-100 tests
    \item VGG19: Layers with the name \textit{features} were filtered, then among 37 \textit{features} layers, 8th, 13th and 17th layers, for all ImageNet and CIFAR-100 tests
    \item ViT: Layers with the name \textit{attn.proj} were filtered, then among 23 \textit{attn.proj} layers, 8th, 13th and 17th layers, for all ImageNet and CIFAR-100 tests
    \item DeiT: Layers with the name \textit{attn.proj} were filtered, then among 24 \textit{attn.proj} layers, 8th, 13th and 17th layers for ImageNet tests, 5th, 6th, 7th layers for CIFAR-100 tests
    \item LeViT: Layers with the name \textit{attn.proj} were filtered, then among 12 \textit{attn.proj} layers, 3th, 5th and 7th layers for ImageNet tests, 5th, 6th, 7th layers for CIFAR-100 tests
\end{enumerate}
Before concatenating the $a_r$ vectors, a specific slicing function for each vector is applied as described in Equation \eqref{eq:slicing}. The specific slicing functions for each corresponding $a_r$ are detailed below: 

\begin{enumerate}
    \item Resnet50: $[:5,:28,:28]$, $[:50,:7,:7]$ and  $[:10,:14,:14]$.
    \item InceptionV3: $[:3,:35,:35]$, $[3:,35:,35]$ and  $[:3,:17,:17]$.
    \item Vgg19: $[:5,:25,:25]$, $[:5,:25,:25]$ and  $[:5,:25,:25]$.
    \item ViT: $[:,:4:200]$, $[:,:4:200]$ and  $[:,:4:,200]$.
    \item DeiT: $[:,:4,:200]$, $[:,:4,:200]$ and  $[:,:4,:200]$.
    \item LeViT $[:4,:14:,14]$, $[:14,:7,:7]$ and  $[:14,:7,:7]$.
\end{enumerate}
The vector $v$ is generated by concatenating the sliced $a_r$ vectors as described in Equation \eqref{eq:v_vector}. After acquiring  $v$ for a model, an MLP with two hidden layers is trained to minimize the MSE loss between $v$ and the feature vector $a_{n-1}$. Adam optimizer with $3 \cdot 10^{-4}$ learning rate is used for the training.

\section{Pseudo-Code for LR}
\label{ap:pseudo}

In Section \ref{sec:lr}, we explain our detection algorithm in detail. Here, in Algorithm \ref{alg:cap}, we also provide a PyTorch \cite{paszke2017automatic} like pseudo code for LR. 

\begin{algorithm}[h]
\caption{Layer Regression (LR)}\label{alg:cap}
\begin{algorithmic}[1]

\Input{ input $x$, selected subset of vector layers $a_r$, slicing functions $s$, DNN  model $g$, feature vector $a_{n-1}$,  LR detector $m$. }  

\Output{ loss $l$ if not training mode.}


\State DNN $g$ takes the input $x$ as $g(x)$
\State Each layer vector $a_r$, process with corresponding slicing function in $s$  
\For{$a_i$ in $a_r$}
        \State $s_r  \gets s_i(a_i(x)) $ 
      \EndFor
\State $v \gets torch.cat(s_r)$
\State feed $v$ into a detection model $m$, such that: $m(v)$
\State calculate the MSE loss between $m(v)$ and $a_{n-1}$:
\State $l \gets MSE(m(v), a_{n-1})$
\If{training}
    \State $l.backward()$
    \State $optimizer.step()$
\Else
    \State Return $l$
\EndIf

\end{algorithmic}
\end{algorithm}

\clearpage

\section{Additional Study: LR in Traffic Sign Detection}
\label{ap:traffic}

\begin{table}[h!]
  \centering
  
  \begin{tabular}{ c |  c  c  c  c | c }
     & FGSM & PGD & Patch & Light & Average\\
    \hline
    LR  & 0.97 & 0.99  & 0.95 & 0.94 & 0.96\\

  \end{tabular}
  \caption{Detection AUROC of LR against four attacks targeting ResNet50 in the traffic sign recognition task.}
  \label{tab:abtraffic}
  
\end{table}

As an additional experiment, we implement LR against attacks that target traffic sign recognition. In this study, ResNet50 is used as target model and attacked with FGSM\citep{fgsm}, PGD\citep{pgd}, Light\citep{light} and Patch \citep{tpatch} attacks. 
In Table \ref{tab:abtraffic}, we show that LR achieves an average AUC score of 96\%, which further proves the applicability and success of our detection method across different domains.

\section{Details of Ablation Study \& Other Domain Experiments}
\label{ap:details_ab}

For Section \ref{sec:ab1}, the experimental settings detailed in \cite{vlad} is followed: "A subset of Kinetics-400 \citep{k400} is
randomly selected for each target model from the videos
that are correctly classified by the respective model. For
each subset, the total number of the videos are between
7700 and 8000 and each class has at least 3, at most 20
instances. An adversarial version of the remaining
20\% portion is generated with each adversarial attack, in a
way that they cannot be correctly classified by the models.
Then the adversarial set is used for evaluation along with the
clean versions."  
Similarly to main experiments, we trained an MLP with 2 hidden layers with the same training hyper-parameters described in  Supplementary \ref{ap:subsets}. For MVIT \citep{mvit} and CSN \citep{csn} models, layers which have the name \textit{conv\_a} and \textit{attn.proj} were filtered respectively. While for both of the models 3th 5th and 7th layers used to form subset $a_r$, for CSN \citep{csn} additional layers 4th, 6th and 8th were also used. $[:, :4, :200]$ and $[:, :4, :3, :7, :7]$ were used as slicing functions for MVIT \citep{mvit} and CSN \citep{csn} respectively.

For Section \ref{sec:ab2}, an MLP with 2 hidden layers with the same training hyper-parameters described in  Supplementary \ref{ap:subsets} is trained. For Wav2vec \citep{w2v} model, layers which have the name \textit{attention.dropout} were filtered, and 3th layer used to form $a_r$. As slicing function $[:, :4, :20, :10]$ is selected. While 500 sound clips from LibriSpeech dataset used for experiments, 2000 sound clips saved for training.

For Supplementary \ref{ap:traffic}, we use the experimental settings as demonstrated in \cite{vilas} and use the  GTSRB \citep{gtsrb}  dataset which includes 43 classes of traffic signs, split into 39,209 176 training images and 12,630 test images. LR detector is trained with the same settings that are specified for Resnet50 in Supplementary \ref{ap:subsets}.

\section{Visualized adversarial examples}

\begin{figure}[h!]
    \centering
    \includegraphics[width=\linewidth]{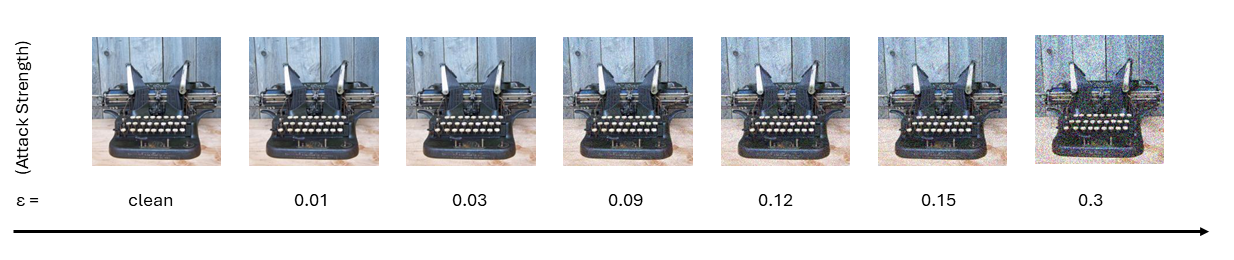}
    \caption{Effects of different attack strength values on a sample. The attack strength parameter $\epsilon$ is increased in the direction of the arrow. In the first sample, there is no attack. Samples are generated with adversarial attack PGD \citep{pgd} and target model ResNet50 \citep{resnet}}.
    \label{fig:apex3}
    \vspace{-3mm}
\end{figure}

\begin{figure}[h]
    \centering
    \includegraphics[width=\linewidth]{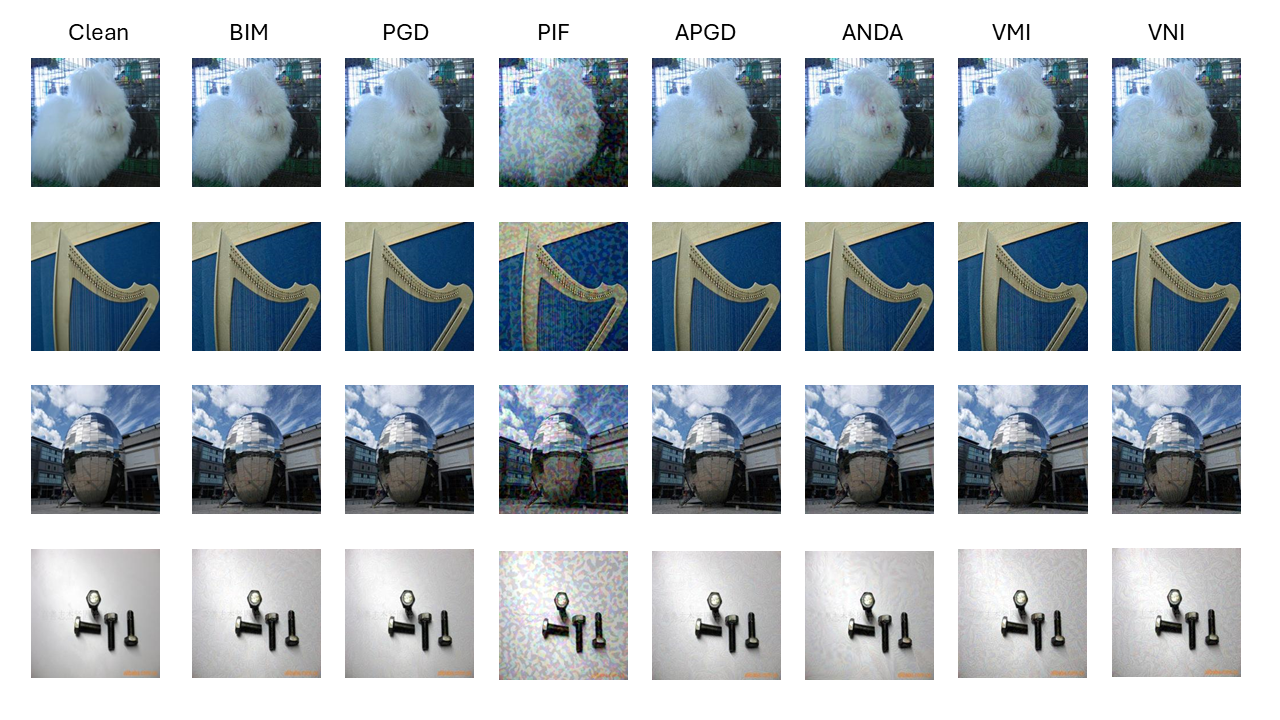}
    \caption{Adversarial examples generated with  BIM \citep{bim}, PGD \citep{pgd}, PIF \citep{pif}, APGD \citep{apgd}, ANDA \citep{anda}, VMI and VNI \citep{vmi_vni} by attacking target model ResNet50 \citep{resnet}. First column represents the clean samples.}
    \label{fig:apex1}
    \vspace{-3mm}
\end{figure}
\vspace{-5mm}
\begin{figure}[h]
    \centering
    \includegraphics[width=\linewidth]{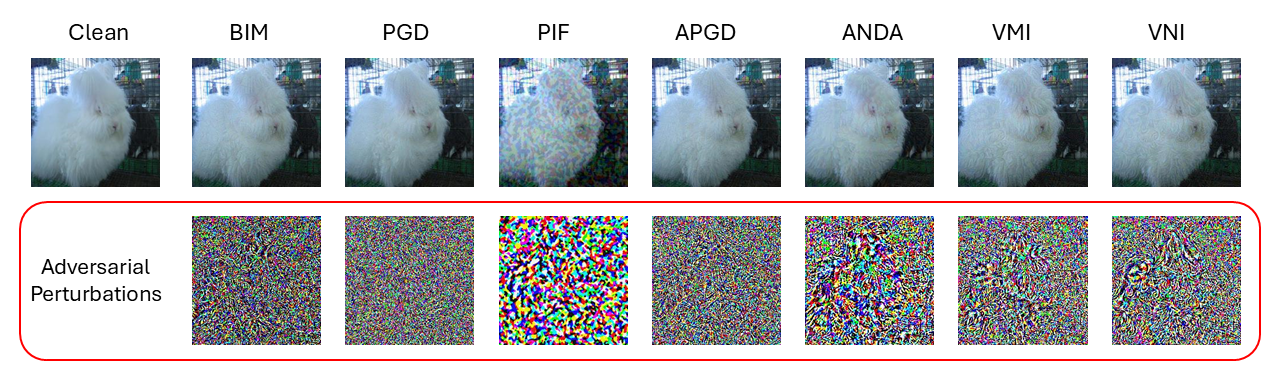}
    \caption{Visualization perturbations generated with BIM \citep{bim}, PGD \citep{pgd}, PIF \citep{pif}, APGD \citep{apgd}, ANDA \citep{anda}, VMI and VNI \citep{vmi_vni} by attacking target model ResNet50 \citep{resnet}. First row shows the clean sample and corresponding {adversarial} example for each attack. Second row demonstrates the noises added to the clean sample by each attack.}
    \label{fig:apex2}
    \vspace{-3mm}
\end{figure}
\vspace{-3mm}

\end{document}